%% file: main_arxiv.tex
\documentclass{article} 
\usepackage{arxiv}

\input{math_commands.tex}

\usepackage[utf8]{inputenc} 
\usepackage[T1]{fontenc}    
\usepackage{hyperref}       
\hypersetup{
    colorlinks=true,
    citecolor=blue
}
\usepackage{url}            
\usepackage{booktabs}       
\usepackage{amsfonts}       
\usepackage{nicefrac}       
\usepackage{microtype}      
\usepackage{lipsum}		
\usepackage{graphicx}
\usepackage{natbib}
\usepackage{doi}

\usepackage[textsize=tiny]{todonotes}

\DeclareUnicodeCharacter{1EF3}{\`y}
\DeclareUnicodeCharacter{003BB}{TT}

\usepackage{xcolor}         
\usepackage[linesnumbered, ruled,vlined]{algorithm2e}
\usepackage{amsmath}
\usepackage{yhmath}
\usepackage{wrapfig}
\usepackage{threeparttable}
\usepackage{multicol}
\usepackage{multirow}
\usepackage{adjustbox}
\usepackage{subfigure}
\usepackage{setspace}
\usepackage{caption}
\usepackage{tikz}
\usetikzlibrary{tikzmark,calc,fit}
\usepackage{listings}

\hypersetup{
colorlinks = true,
urlcolor=blue
}

\definecolor{codegreen}{rgb}{0,0.6,0}
\definecolor{codegray}{rgb}{0.5,0.5,0.5}
\definecolor{codepurple}{rgb}{0.58,0,0.82}
\definecolor{backcolour}{rgb}{0.95,0.95,0.92}
\lstdefinestyle{mystyle}{
    inputencoding=utf8,
    extendedchars=true,
    commentstyle=\color{codegreen},
    keywordstyle=\color{magenta},
    numberstyle=\tiny\color{codegray},
    stringstyle=\color{codepurple},
    basicstyle=\ttfamily\small,
    breakatwhitespace=false,         
    breaklines=true,                 
    captionpos=b,                    
    keepspaces=true,                 
    numbers=left,                    
    numbersep=5pt,                  
    showspaces=false,                
    showstringspaces=false,
    showtabs=false,                  
    tabsize=4,
    literate={λ}{{$\lambda$}}1
}
\SetCommentSty{mycommfont}
\definecolor{Gray}{rgb}{0.85, 0.85, 0.85}
\definecolor{lightGray}{rgb}{0.93, 0.93, 0.93}

\newcommand{\Tau}{\mathcal{T}}
\newcommand{\D}{\mathcal{D}}

\title{SSFL: Tackling Label Deficiency in Federated Learning via Personalized Self-Supervision}

\author{Chaoyang He, Zhengyu Yang, Erum Mushtaq, Sunwoo Lee\\ \textbf{Mahdi Soltanolkotabi, Salman Avestimehr}\\
Viterbi School of Engineering \\
University of Southern California \\
\texttt{\{chaoyang.he,yang765,emushtaq,sunwool,soltanol,avestime\}@usc.edu}
}

\begin{document}
\maketitle
\begin{abstract}
Federated Learning (FL) is transforming the ML training ecosystem from a centralized over-the-cloud setting to distributed training over edge devices in order to strengthen data privacy, reduce data migration costs, and break regulatory restrictions. An essential, but rarely studied, challenge in FL is \textit{label deficiency} at the edge. This problem is even more pronounced in FL, compared to centralized training, due to the fact that FL users are often reluctant to label their private data and edge devices do not provide an ideal interface to assist with annotation. Addressing label deficiency is also further complicated in FL, due to the heterogeneous nature of the data at edge devices and the need for developing personalized models for each user.
We propose a self-supervised and personalized federated learning framework, named (\texttt{SSFL}), and a series of algorithms under this framework which work towards addressing these challenges. First, under the \texttt{SSFL} framework, we analyze the compatibility of various centralized self-supervised learning methods in FL setting and demonstrate that \texttt{SimSiam} networks performs the best with the standard \texttt{FedAvg} algorithm. Moreover, to address the data heterogeneity at the edge devices in this framework, we have innovated a series of algorithms that broaden existing supervised personalization algorithms into the setting of self-supervised learning including \texttt{perFedAvg}, \texttt{Ditto}, and local fine-tuning, among others. We further propose a novel personalized federated self-supervised learning algorithm, \texttt{Per-SSFL}, which balances personalization and consensus by carefully regulating the distance between the local and global representations of data. To provide a comprehensive comparative analysis of all proposed algorithms, we also develop a distributed training system and related evaluation protocol for \texttt{SSFL}. Using this training system, we conduct experiments on a synthetic non-I.I.D. dataset based on CIFAR-10, and an intrinsically non-I.I.D. dataset GLD-23K. Our findings show that the gap of evaluation accuracy between supervised learning and unsupervised learning in FL is both small and reasonable. The performance comparison indicates that representation regularization-based personalization method is able to outperform other variants. Ablation studies on SSFL are also conducted to understand the role of batch size, non-I.I.D.ness, and the evaluation protocol.
\end{abstract}

\section{Introduction}
Federated Learning (FL) is a contemporary distributed machine learning paradigm that aims at strengthening data privacy, reducing data migration costs, and breaking regulatory restrictions \citep{Kairouz2021AdvancesAO,Wang2021AFG}. It has been widely applied to computer vision, natural language processing, and data mining. 
However, there are two main challenges impeding its wider adoption in machine learning. One is data heterogeneity, which is a natural property of FL in which diverse clients may generate datasets with different distributions due to behavior preferences (e.g., the most common cause of heterogeneity is skewed label distribution which might result from instances where some smartphone users take more landscape pictures, while others take more photos of daily life). The second challenge is label deficiency at the edge, which is relatively less studied. This issue is more severe at the edge than in a centralized setting because users are reluctant to annotate their private and sensitive data, and/or smartphones and IoT devices do not have a user-friendly interface to assist with annotation.

\begin{figure}[ht!]
    \centering
    \includegraphics[width=1.0\textwidth]{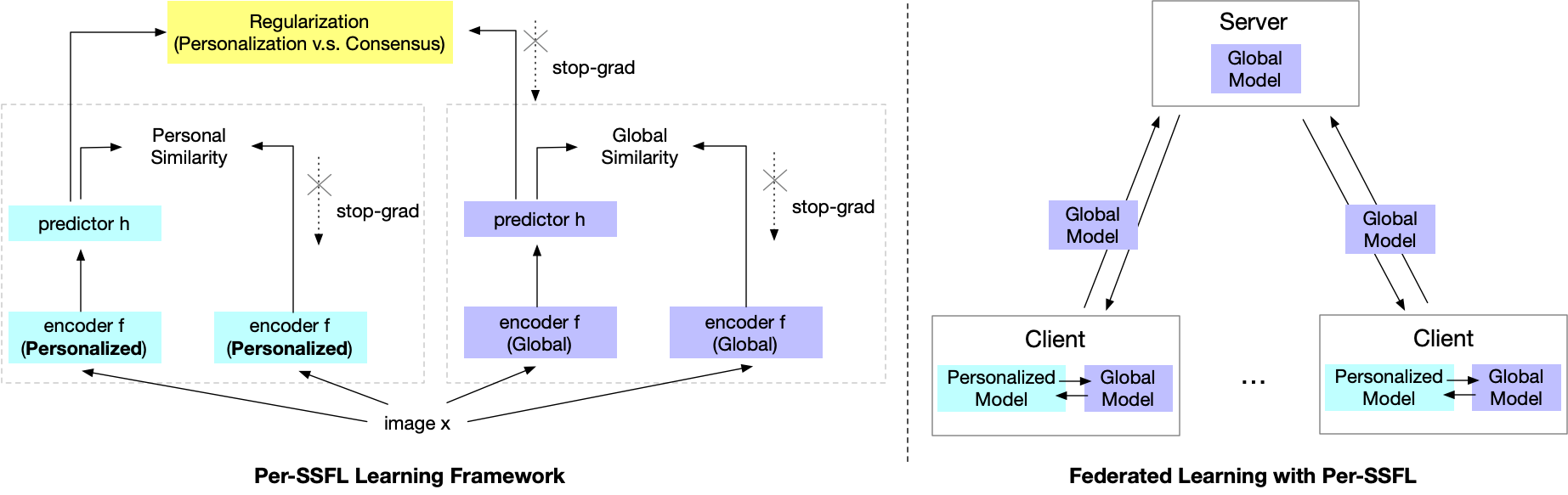}
    \caption{Depiction of the Self-supervised and Personalized Federated Learning (SSFL) framework.}
    \label{fig:ssl_idea}
\end{figure}

To mitigate the data heterogeneity issue among clients, researchers have proposed algorithms for training a global model \texttt{FedAvg} \citep{mcmahan2017communication}, \texttt{FedProx} \citep{li2018federated}, \texttt{FedNova} \citep{wang2020tackling}, \texttt{FedOPT} \citep{reddi2020adaptive}, as well as personalized FL frameworks (e.g., \texttt{pFedMe}, \texttt{Ditto}, \texttt{Per-FedAvg}). These algorithms all depend on the strong assumption that the data at the edge has sufficient labels. 
To address the label deficiency issue in FL, recent works  \citep{liu2020rc,long2020fedsemi,itahara2020distillation,jeong2020federated,liang2021self,zhao2020semi,zhang2020federated,zhang2020improving} assume that the server or client has a fraction of labeled data and use semi-supervised methods such as consistency loss \citep{miyato2018virtual} or pseudo labeling \citep{Lee2013PseudoLabelT} to train a global model. 
A more realistic but challenging setting is fully unsupervised training. Although a recent work in FL \citep{saeed2020federated} attempts to address this challenge through Siamese networks proposed around thirty years ago \citep{bromley1993signature}, its design does not tackle data heterogeneity for learning personalized models, and it only trains on small-scale sensor data in IoT devices. Moreover,
these existing works in FL have not examined recent progress in the Self-Supervised Learning (SSL) community where methods such as \texttt{SimCLR} \citep{chen2020simple}, \texttt{SwAV}\citep{caron2021unsupervised}, \texttt{BYOL} \citep{grill2020bootstrap}, and \texttt{SimSiam} \citep{chen2020exploring} have shown tremendous improvement in reducing the amount of labeled data required to achieve state-of-the-art performance. As such, it remains still unclear how these SSL methods can be incorporated into FL and how well they would perform, especially when intertwined with the data heterogeneity challenge that does not exist in centralized training.

In this paper, we propose Self-Supervised Federated Learning (\texttt{SSFL}), a unified self-supervised and personalized federated learning framework, and a series of algorithms under this framework to address these challenges. As shown in Figure \ref{fig:ssl_idea}, this framework brings state-of-the-art self-supervised learning algorithms to the realm of FL in order to enable training without using any supervision, while also integrating model personalization to deal with data heterogeneity (Section \ref{sec:general_framework}). More specifically, under the \texttt{SSFL} framework, we analyze the compatibility of various centralized self-supervised learning methods in the FL setting and demonstrate that \texttt{SimSiam} networks performs the best with the standard FedAvg algorithm (Section \ref{sec:global-ssff-32}). Moreover, to address the data heterogeneity at edge devices, we have innovated a series of algorithms that broaden the reach of existing supervised personalization algorithms into the setting of self-supervised learning, including  \texttt{perFedAvg} \citep{fallah2020personalized}, \texttt{Ditto} \citep{li_ditto_2021}, and \texttt{local fine-tuning}, among others. We further propose a novel personalized federated self-supervised learning algorithm, \texttt{per-SSFL} (Section \ref{sec:per-SSFL}), which balances personalization and consensus by carefully regulating the distance between the local and global representations of data (shown as the yellow block in Figure \ref{fig:ssl_idea}). 

To provide a comprehensive and comparative analysis of the proposed algorithms, we also develop a distributed training system and evaluation protocol for SSFL. Using this training system, we conduct experiments on a synthetic non-I.I.D. dataset based on CIFAR-10 and a natural non-I.I.D. dataset GLD-23K. Our experimental results demonstrate that all algorithms in our framework work reliably. In FL, the gap of evaluation accuracy between supervised learning and unsupervised learning is small. Personalized \texttt{SSFL} performs better than \texttt{FedAvg}-based \texttt{SSFL}. We also conduct ablation studies to fully understand the \texttt{SSFL} framework, namely the role of batch size, different degrees of non-I.I.D.ness, and performance in more datasets. Finally, our unified API design can serve as a suitable platform and baseline, enabling further developments of more advanced \texttt{SSFL} algorithms.

\section{Preliminaries}
SSFL builds upon two fundamental areas in machine learning: federated optimization and self-supervised learning. Thus, we first introduce some basics and formulations in these areas.

\subsection{Federated Optimization}
Federated optimization refers to the distributed optimization paradigm that a network of $K$ devices collaboratively solve a machine learning task. In general, it can be formulated as a distributed optimization problem with the form \citep{mcmahan2017communication}: $
\min_{\theta} \sum_{k=1}^{K} \frac{|D_k|}{|D|} \mathcal{L}(\theta,D_k) $.
Here,  each device $k$ has a local dataset $D_k$ drawn from a local distribution $X_k$. The combined dataset $D=\cup_{k=1}^K D_k$ is the union of all local datasets $D_k$. $\theta$ represents the model weight of a client model. $\mathcal{L}$ is the client's local loss function that measures the local empirical risk over the heterogeneous dataset $\mathcal{D}^k$. Under this formulation, to mitigate the non-I.I.D. issue, researchers have proposed algorithms such as \texttt{FedProx} \citep{li2018federated}, \texttt{FedNova} \citep{wang2020tackling}, and \texttt{FedOPT} \citep{reddi2020adaptive} for training a global model, as well as personalized FL frameworks such as \texttt{Ditto} \citep{li_ditto_2021}, and \texttt{Per-FedAvg} \citep{fallah_personalized_2020-2}. All of these algorithms have a strong assumption that data at the edge have sufficient labels, meaning that their analysis and experimental settings are based on a supervised loss function, such as the cross-entropy loss for image classification. 

\subsection{Self-supervised Learning}
\label{sec:ssfl}

Self-supervised learning (SSL) aims to learn meaningful representations of samples without human supervision. 
Formally, it aims to learn an encoder function $f_\theta: \mathcal{X} \mapsto \mathbb{R}^d$ where $\theta$ is the parameter of the function, $\mathcal{X}$ is the unlabeled sample space (e.g. image, text), and the output is a $d$ dimensional vector containing enough information for downstream tasks such as image classification and segmentation. The key to SSL's recent success is the inductive bias that ensures a good representation encoder remains consistent under different perturbations of the input (i.e. consistency regularization). 

One prominent example among recent advances in modern SSL frameworks is the Siamese architecture \citep{bromley1993signature} and its improved variants \texttt{SimCLR} \citep{chen2020simple}, \texttt{SwAV} \citep{caron2021unsupervised}, \texttt{BYOL} \citep{grill2020bootstrap}, 
and \texttt{SimSiam} \citep{chen2020exploring}. Here we review the most elegant architecture, \texttt{SimSiam}, and defer the description and comparison of the other three to Appendix \ref{app:ssl}. \texttt{SimSiam} proposes a two-head architecture in which two different views (augmentations) of the same image are encoded by the same network $f_\theta$. Subsequently, a predictor Multi Layer Perceptron (MLP) $h_\theta$ and a \textit{stop-gradient} operation denoted by $\widehat{\cdot}$ are applied to both heads. In the SSL context, ``stop gradient'' means that the optimizer stops at a specific neural network layer during the back propagation and views the parameters in preceding layers as constants. Here, $\theta$ is the concatenation of the parameters of the encoder network and the predictor MLP. The algorithm aims to minimize the negative cosine similarity $\mathcal{D}(\cdot, \cdot)$ between two heads. More concretely, the loss is defined as
\begin{equation}
    \mathcal{L}_{\mathrm{SS}}(\theta,D)= \frac{1}{|D|} \sum_{x\in D}\mathcal{D}(f_{\theta}(\Tau(x)),\widehat{h_{\theta}(f_{\theta}(\Tau(x)))}),
\label{eq:global_ssl_opt}
\end{equation}
where $\Tau$ represents stochastic data augmentation and $D$ is the data set. 



\section{SSFL: Self-supervised Federated Learning}

In this section, we propose \texttt{SSFL}, a unified framework for self-supervised federated learning. Specifically, we introduce the method by which \texttt{SSFL} works for collaborative training of a global model and personalized models, respectively.

\subsection{General Formulation}
\label{sec:general_framework}

We formulate self-supervised federated learning as the following distributed optimization problem:
\begin{equation}
    \min_{\substack{\Theta\\ \{\theta_k\}_{k \in [K]}}} G \left(\mathcal{L}\left(\theta_{1}, \Theta; X_{1}\right), \dots, \mathcal{L}\left(\theta_{K}, \Theta; X_{K}\right) \right)
    \label{eq:ssfl_target}
\end{equation}
where $\theta_k$ is the parameter for the local model $(f_{\theta_k}, h_{\theta_k})$; $\Theta$ is the parameter for the global model $(f_{\Theta}, h_\Theta)$; $\mathcal{L}(\theta_k, \Theta; X_{k})$ is a loss measuring the quality of representations encoded by $f_{\theta_k}$ and $f_{\Theta}$ on the local distribution $X_k$; and $G(\cdot)$ denotes the aggregation function (e.g. sum of client losses weighted by $\frac{|D_k|}{|D|}$). To capture the two key challenges in federated learning (data heterogeneity and label deficiency), we hold two core assumptions in the proposed framework: (1) $X_{k}$ of all clients are heterogeneous (non-I.I.D.), and (2) there is no label.

To tackle the above problem, we propose a unified training framework for federated self-supervised learning, as described in Algorithm \ref{alg:ssfl}. This framework can handle both non-personalized and personalized federated training. In particular, if one enforces the constraint $\theta_k = \Theta$ for all clients $k\in[K]$, the problem reduces to learning a global model. When this constraint is not enforced, $\theta_k$ can be different for each client, allowing for model personalization. \texttt{ClientSSLOPT} is the local optimizer at the client side which solves the local sub-problem in a self-supervised manner. \texttt{ServerOPT} takes the update from the client side and generates a new global model for all clients.

\newcommand{\clientOpt}{\colorbox{blue!18}{\textsc{ClientSSLOpt}}}
\newcommand{\serverOpt}{\colorbox{green!18}{\textsc{ServerOpt}}}
\begin{center}
\scalebox{0.85}{
\begin{minipage}{\linewidth}
\begin{algorithm}[H]
\small
\SetKwInOut{Input}{input}\SetKwInOut{Output}{return}
\Input{$K, T, \Theta^{(0)}, \{\theta_k^{(0)}\}_{k\in[K]}, \clientOpt, \serverOpt$}
\For{$t=0,\ldots,T-1$}{
    Server randomly selects a subset of devices $S^{(t)}$\\
    Server sends the current global model $\Theta^{(t)}$ to $S^{(t)}$\\
    \For{device $k \in S^{(t)}$ in parallel}{
        Solve local sub-problem of equation \ref{eq:ssfl_target}:
        \[
            \theta_k, \Theta_k^{(t)} \leftarrow \clientOpt (\theta_k^{(t)}, \Theta^{(t)}, \nabla \mathcal{L}\left(\theta_{k}, \Theta; X_{k}\right))
        \]\\
        Send $\Delta_k^{(t)}:=\Theta_k^{(t)}-\Theta^{(t)}$ back to server\\
    }
    $\Theta^{(t+1)} \leftarrow \serverOpt \left( \Theta^{(t)}, \{ \Delta_k^{(t)} \}_{k \in S^{(t)}} \right)$
}
\Output{$\{\theta_k\}_{k\in[K]}, \Theta^{(T)}$}
\caption{SSFL: A Unified Framework for Self-supervised Federated Learning}
\label{alg:ssfl}
\end{algorithm}
\end{minipage}%
}
\end{center}

Next, we will introduce specific forms of \texttt{ClientSSLOPT} and \texttt{ServerOPT} for global training and personalized training.

\subsection{Global-SSFL: Collaboration Towards a Global Model without Supervision}
\label{sec:global-ssff-32}

To train a global model using \texttt{SSFL}, we design a specific form of \texttt{ClientSSLOPT} using \texttt{SimSiam}. We choose \texttt{SimSiam} over other contemporary self-supervised learning frameworks (e.g., \texttt{SimSiam}, \texttt{SwAV}, \texttt{BYOL}) based on the following analysis as well as experimental results (see Section \ref{sec:exp_compare_simsaim}).

\textbf{The simplicity in neural architecture and training method.} \texttt{SimSiam}'s architecture and training method are relatively simple. For instance, compared with \texttt{SimCLR}, \texttt{SimSiam} has a simpler loss function; compared with \texttt{SwAV}, \texttt{SimSiam} does not require an additional neural component (prototype vectors) and Sinkhorn-Knopp algorithm; compared with \texttt{BYOL}, \texttt{SimSiam} does not need to maintain an additional moving averaging network for an online network. Moreover, the required batch size of \texttt{SimSiam} is the smallest, making it relatively friendly for resource-constrained federated learning. A more comprehensive comparison can be found in Appendix \ref{app:ssl}.

\textbf{Interpretability of SimSiam leads to simpler local optimization.} More importantly, \texttt{SimSiam} is more interpretable from an optimization standpoint which simplifieds the local optimization. In particular, it can be viewed as an implementation of an Expectation-Maximization (EM) like algorithm, meaning that optimizing $\mathcal{L}_\mathrm{SS}$ in Equation \ref{eq:global_ssl_opt} is implicitly optimizing the following objective
\begin{equation}
    \min_{\theta, \eta} \E_{\substack{\Tau\\x \sim X}}\left[\left\|f_{\theta}(\Tau(x))-\eta_x\right\|_{2}^{2}\right].
\label{eq:global_e}
\end{equation}
Here, $f_{\theta}$ is the encoder neural network parameterized by $\theta$. $\eta$ is an extra set of parameters, whose size is proportional to the number of images, and $\eta_x$ refers to using the image index of $x$ to access a sub-vector of $\eta$. This formulation is w.r.t. both $\theta$ and $\eta$ and can be optimized via an alternating algorithm. At time step $t$, the $\eta^{t}_x$ update takes the form $\eta^{t}_x\leftarrow \E_{\Tau}\left[f_{\theta^{t}}(\Tau(x))\right]$, indicating that $\eta^{t}_x$ is assigned the average representation of $x$ over the distribution of augmentation. However, it is impossible to compute this step by going over the entire dataset during training. Thus, \texttt{SimSiam} uses one-step optimization to approximate the EM-like two-step iteration by introducing the predictor $h_\theta$ to approximate $\eta$ and learn the expectation (i.e. $h_\theta\left(z\right) \approx \mathbb{E}_{\mathcal{T}}[f_\theta(\mathcal{T}(x))]$) for any image $x$. After this approximation, the expectation $\mathbb{E}_{\mathcal{T}}[\cdot]$ is ignored because the sampling of $\mathcal{T}$ is implicitly distributed across multiple epochs. Finally, we can obtain the self-supervised loss function in Equation \ref{eq:global_ssl_opt}, in which negative cosine similarity $\mathcal{D}$ is used in practice (the equivalent $L_2$ distance is used in Equation \ref{eq:global_e} for the sake of analysis). Applying equation \ref{eq:global_ssl_opt} as \texttt{ClientSSLOPT} simplifies the local optimization for each client in a self-supervised manner. 

\subsection{Per-SSFL: Learning Personalized Models without Supervision}
\label{sec:per-SSFL}

In this section, we explain how SSFL addresses the data heterogeneity challenge by learning personalized models.
Inspired by the interpretation in Section \ref{sec:global-ssff-32}, we define the following sub-problem for each client $k \in [K]$:
\begin{equation}
    \label{eq:perssfl1}
    \begin{aligned}
    \min_{\theta_{k},\eta_{k}} & \quad\E_{\substack{\Tau\\
    x\sim X_{k}
    }
    }\left[\left\Vert f_{\theta_{k}}(\Tau(x))-\eta_{k,x}\right\Vert _{2}^{2}+\frac{\lambda}{2}\left\Vert \eta_{k,x}-\mathcal{H}_{x}^{*}\right\Vert _{2}^{2}\right]\\
    \text{s.t.} & \quad\Theta^{*},\mathcal{H}^{*}\in\arg\min_{\Theta,\mathcal{H}}\sum_{i=1}^{n} \frac{|D_k|}{|D|} \E_{\substack{\Tau\\
    x\sim X_{i}
    }
    }\left[\left\Vert f_{\Theta}(\Tau(x))-\mathcal{H}_{x}\right\Vert _{2}^{2}\right]
    \end{aligned}
\end{equation}

Compared to global training, we additionally include $\Theta$, the global model parameter, and $\mathcal{H}$, the global version of $\eta$, and the expected representations which correspond to the personalized parameters $\theta_k$ and $\eta_{k}$. In particular, through the term $\left\Vert \eta_{k,x}-\mathcal{H}_{x}^{*}\right\Vert _{2}^{2}$, we aim for the expected local representation of any image $x$ to reside within a neighborhood around the expected global representation of $x$. Therefore, by controlling the radius of the neighborhood, hyperparameter $\lambda$ helps to balance consensus and personalization.

\input{listing/PerFedSimSiam}

We see that Equation \ref{eq:perssfl1} in the above objective is an optimization problem w.r.t. both $\theta$ and $\eta$. However, as the above target is intractable in practice, following an analysis similar to Section \ref{sec:global-ssff-32}, we use the target below as a surrogate:
\begin{align}
\min_{\theta_{k}}&\quad\mathcal{L}_{\mathrm{SS}}\left(\theta_{k},D_{k}\right)+\frac{\lambda}{\left|D_{k}\right|}\sum_{x\in D_{k}}\mathcal{D}\left(h_{\theta_{k}}\left(f_{\theta_{k}}\left(\Tau(x)\right)\right),h_{\Theta^{*}}\left(f_{\Theta^{*}}\left(\Tau(x)\right)\right)\right)\\\text{s.t.}&\quad\Theta^{*}\in\arg\min_{\Theta}\mathcal{L}_{\mathrm{SS}}\left(\Theta,D\right)
\end{align}
In practice, $\Theta$ can be optimized independently of $\theta^k$ through the \texttt{FedAvg} \citep{mcmahan2017communication} algorithm. To make the computation more efficient, we also apply the symmetrization trick proposed in \citep{chen2020exploring}. We refer to this algorithm as \texttt{Per-SSFL} and provide a detailed description in Algorithm \ref{alg:pfss} (also illustrated in Fig. \ref{fig:ssl_idea}).

\textbf{Regarding the theoretical analysis.} To our knowledge, all self-supervised learning frameworks do not have any theoretical analysis yet, particularly the SimSiam dual neural network architecture. Our formulation and optimization framework are interpretable, they are built based on an EM-like algorithm for SimSiam and minimizing the distance between the private model and the global model’s data representation.

\paragraph{Innovating baselines to verify SSFL.} \textit{Note that we have not found any related works that explore a Siamese-like SSL architecture in an FL setting.} As such, to investigate the performance of our proposed algorithm, we further propose several other algorithms that can leverage the \texttt{SSFL} framework. 1.
\texttt{LA-SSFL}. We apply \texttt{FedAvg} \citep{mcmahan2017communication} on the \texttt{SimSiam} loss $\mathcal{L}_\mathrm{SS}$ for each client to obtain a global model. We perform one step of SGD on the clients' local data for local adaption; 2. \texttt{MAML-SSFL}. This algorithm is inspired by perFedAvg \citep{fallah_personalized_2020-2} and views the personalization on each devices as the inner loop of MAML \citep{finn_model-agnostic_2017}. It aims to learn an encoder that can be easily adapted to the clients' local distribution. During inference, we perform one step of SGD on the global model for personalization; 3. \texttt{BiLevel-SSFL}. Inspired by Ditto \citep{li_ditto_2021}, we learn personalized encoders on each client by restricting the parameters of all personalized encoders to be close to a global encoder independently learned by weighted aggregation. More details of these algorithms, formulation, and pseudo code are introduced in Appendix \ref{app:algorithms}. In Section \ref{sec:exp-per-4-2}, we will show the comparison results for these proposed \texttt{SSFL} algorithmic variants.

\section{Training System and Evaluation Pipeline for SSFL}
\label{sec:protocol}

\textbf{A Distributed Training System to accelerate the algorithmic exploration in SSFL framework.} We also contributed to reproducible research via our distributed training system. This is necessary for two reasons: (1) Running a stand-alone simulation (training client by client sequentially) like most existing FL works requires a prohibitively long training time when training a large number of clients. In SSFL, the \textit{model size} (e.g., ResNet-18 v.s. shallow CNNs used in the original FedAvg paper) and the \textit{round number} for convergence (e.g., 800 epochs in the centralized SimSiam framework) is relatively larger than in FL literature. By running all clients in parallel on multiple CPUs/GPUs, we can largely accelerate the process. (2) Given that SSFL is a unified and generic learning framework, researchers may develop more advanced ways to improve our work. As such, we believe it is necessary to design unified APIs and system frameworks in line with the algorithmic aspect of SSFL. See  Appendix \ref{app:system} for more details on our distributed training system.

\paragraph{Evaluation Pipeline.} In the training phase, we use a \texttt{KNN} classifier \citep{wu2018unsupervised} as an online indicator to monitor the quality of the representations generated by the \texttt{SimSiam} encoder. For \texttt{Global-SSFL}, we report the \texttt{KNN} test accuracy using the global model and the global test data, while in \texttt{Per-SSFL}, we evaluate all clients' local encoders separately with their local test data and report their averaged accuracy.
After self-supervised training, to evaluate the performance of the trained encoder, we freeze the encoder and attach a linear classifier to the output of the encoder.
For \texttt{Global-SSFL}, we can easily verify the performance of \texttt{SimSiam} encoder by training the attached linear classifier with \texttt{FedAvg}. However, for \texttt{Per-SSFL}, each client learns a personalized \texttt{SimSiam} encoder. 
As the representations encoded by personalized encoders might reside in different spaces, using a single linear classifier trained by \texttt{FedAvg} to evaluate these representations is unreasonable (see experiments in Section \ref{sec:exp-eval-protocol}). As such, we suggest an additional evaluation step to provide a more representative evaluation of \texttt{Per-SSFL}'s performance: for each personalization encoder, we use the entire training data to train the linear classifier but evaluate on each client's local test data.


\section{Experiments}
\label{sec:exp}
In this section, we introduce experimental results for SSFL with and without personalization and present a performance analysis on a wide range of aspects including the role of batch size, different degrees of non-IIDness, and understanding the evaluation protocol.  

\textbf{Implementation.} We develop the SSFL training system to simplify and unify the algorithmic exploration. Details of the training system design can be found in Appendix \ref{app:system}. We deploy the system in a distributed computing environment which has 8 NVIDIA A100-SXM4 GPUs with sufficient memory (40 GB/GPU) to explore different batch sizes (Section \ref{sec:batch_size}). Our training framework can run multiple parallel training workers in a single GPU, so it supports federated training with a large number of clients. The client number selected per round used in all experiments is $10$, which is a reasonable setting suggested by recent literature \citep{reddi2020adaptive}. 

\textbf{Learning Task.} Following \texttt{SimCLR} \citep{chen2020simple}, \texttt{SimSiam} \citep{chen2020exploring}, \texttt{BYOL} \citep{grill2020bootstrap}, and \texttt{SwAV} \citep{caron2020unsupervised} in the centralized setting, we evaluate SSL for the \textit{image classification} task and use representative datasets for federated learning. 

\textbf{Dataset.} We run experiments on synthetic non-I.I.D. dataset CIFAR-10 and intrinsically non-I.I.D. dataset Google Landmark-23K (GLD-23K), which are suggested by multiple canonical works in the FL community \citep{reddi2020adaptive,chaoyanghe2020fedml,kairouz2019advances}. For the non-I.I.D. setting, we distribute the dataset using a Dirichlet distribution \citep{hsu2019measuring}, which samples $\mathbf{p}_{c} \sim \operatorname{Dir}(\alpha)$ (we assume a uniform prior distribution) and allocates a $\mathbf{p}_{c, k}$ proportion of the training samples of class $c$ to local client $k$. 
We provide a visualization of the data distribution in Appendix \ref{app:data}.

\textbf{Model Architecture.} For the model architecture, ResNet-18 is used as the backbone of the \texttt{SimSiam} framework, and the predictor is the same as that in the original paper.

Next, we focus on results from the curated CIFAR-10 dataset and defer GLD-23K to Appendix \ref{app:gld23k}. 

\subsection{Comparisons on SimSiam, SimCLR, SwAV, and BYOL}
\label{sec:exp_compare_simsaim}

Our first experiment determines which SSL method is ideal for FL settings. We run experiments using FedAvg for these four methods and obtain two findings: (1) \texttt{SimSiam} outperforms \texttt{SimCLR} in terms of accuracy; (2) \texttt{BYOL} and \texttt{SwAV} do not work in FL; we tested a wide range of hyper-parameters, but they still are unable to converge to a reasonable accuracy. These experimental results confirm our analysis in Section \ref{sec:global-ssff-32}.

\subsection{Evaluation on Global-SSFL}
\label{sec:exp-ssfl}
\begin{figure}[h!]
\centering
\subfigure[\label{fig:1A} Training Time Accuracy]
{{\includegraphics[width=0.45\textwidth]{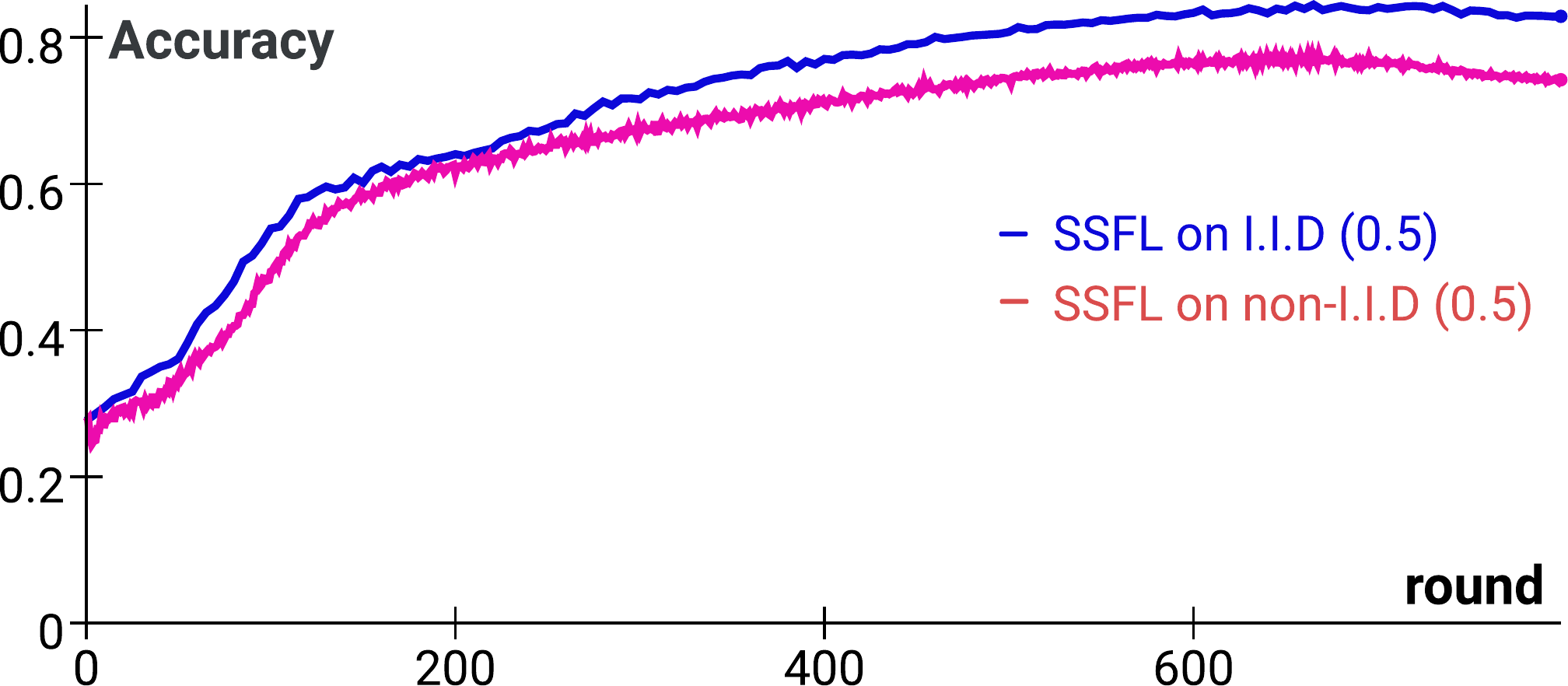}}}
\hspace{0.4cm}
\subfigure[\label{fig:1B} Training Loss]
{{\includegraphics[width=0.45\textwidth]{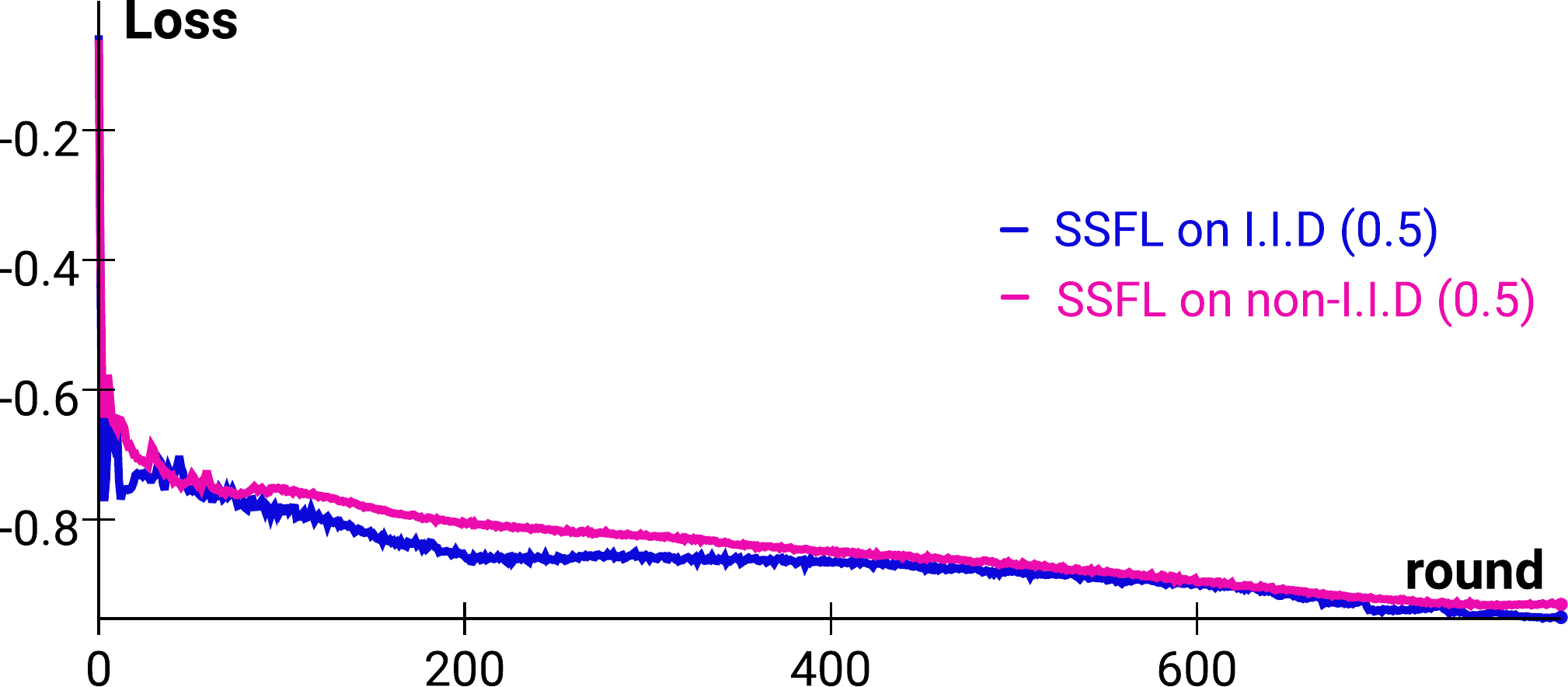}}}
\caption{Training and Evaluation using SSFL}
\label{fig:SSFL_without_per}
\end{figure}
The goal of this experiment is to understand the accuracy gap between supervised and self-supervised federated learning in both I.I.D. and non-I.I.D. settings where we aim to train a global model from private data from clients.

\textbf{Setup and Hyper-parameters.} We evaluate Global-SSFL using non-I.I.D. data from CIFAR-10: we set $\alpha=0.1$ for the Dirichlet distribution. For supervised learning, the test accuracy is evaluated on a global test dataset. For self-supervised training, we follow the evaluation protocol introduced in Section \ref{sec:protocol}.
We use SGD with Momentum as the client-side optimizer and a learning rate scheduler across communication rounds. We searched the learning rate on the grid of $\{0.1, 0.3, 0.01, 0.03\}$ and report the best result. The experiment is run three times using the same learning rate with fixed random seeds to ensure reproducibility. The training lasts for 800 rounds, which is sufficient to achieve convergence for all methods. More hyperparameters are in Appendix \ref{app:hps}. 

We display the training curves in Figure \ref{fig:SSFL_without_per} which demonstrates that \texttt{SSFL} can converge reliably in both I.I.D. and non-I.I.D. settings. 
For the I.I.D. data, we find that \texttt{SSFL} can achieve the same accuracy as the centralized accuracy report in the \texttt{SimSiam} paper \citep{chen2020exploring}. For the non-I.I.D. data, SSFL achieves a reasonable accuracy compared to the centralized accuracy. The accuracy comparisons in different dimensions (supervised v.s. self-supervised; I.I.D. v.s. non-I.I.D.) are summarized in Table \ref{table1}.

\begin{minipage}{\textwidth}
  \begin{minipage}[b]{0.4\textwidth}
    \centering
    \resizebox{\columnwidth}{!}{
   \begin{tabular}{cccc}
    \toprule
    & \multicolumn{2}{c}{Accuracy} & \\
    \cmidrule(lr){2-3}
       & Supervised & Self-Supervised & Acc. Gap \\
       \midrule
       I.I.D  & 0.932 & 0.914 & 0.018 \\
       non-I.I.D  & 0.8812 & 0.847 & 0.0342 \\
       Acc. Gap & 0.0508 & 0.06 & N/A\\
       \bottomrule
       \end{tabular}}
\captionof{table}{Evaluation accuracy comparison between supervised FL and \texttt{SSFL}.}
\label{table1}
  \end{minipage}
  \hfill
  \begin{minipage}[b]{0.41\textwidth}
    \centering
    \resizebox{\columnwidth}{!}{
  \begin{tabular}{ccc}
\toprule
Method & KNN Indicator & Evaluation \\
\midrule
\texttt{LA-SSFL} & 0.9217 & 0.8013\\
\texttt{MAML-SSFL} & 0.9355 & 0.8235\\
\texttt{BiLevel-SSFL} & 0.9304 & 0.8137\\
\texttt{Per-SSFL} & 0.9388 & 0.8310\\
\bottomrule    
    \end{tabular}}
    \setlength{\belowcaptionskip}{-0.2cm}
      \captionof{table}{Evaluation Accuracy for Various Per-SSFL Methods.}
      \label{table:per-ssfl}
    \end{minipage}
  \end{minipage}

\subsection{Evaluation on Per-SSFL}
\label{sec:exp-per-4-2}

Based on the results of SSFL with \texttt{FedAvg}, we further add the personalization components for SSFL introduced in Section \ref{sec:per-SSFL} (Per-SSFL).

\textbf{Setup and Hyper-parameters.} For a fair comparison, we evaluate \texttt{Per-SSFL} on non-I.I.D. data from CIFAR-10 and set $\alpha=0.1$ for the Dirichlet distribution. For \texttt{Per-SSFL} training, we follow the evaluation protocol introduced in Section \ref{sec:protocol}. Similar to SSFL, we use SGD with Momentum as the client-side optimizer and the learning rate scheduler across communication rounds. We search for the learning rate on a grid of $\{0.1, 0.3, 0.01, 0.03\}$ and report the best result. For \texttt{Per-SSFL} and \texttt{BiLevel-SSFL}, we also tune the $\lambda$ of the regularization term with a search space $\{1, 10, 0.1, 0.01, 0.001\}$.  The experiments are run three times with the same learning rate and with fixed random seeds to ensure reproducibility. The training also lasts for 800 communication rounds, which is the same as \texttt{Global-SSFL}. Other hyperparameters can be found in Appendix \ref{app:hps}. 

\begin{figure}[h!]
\centering
\subfigure[\label{fig:1A} Training Loss]
{{\includegraphics[width=0.45\textwidth]{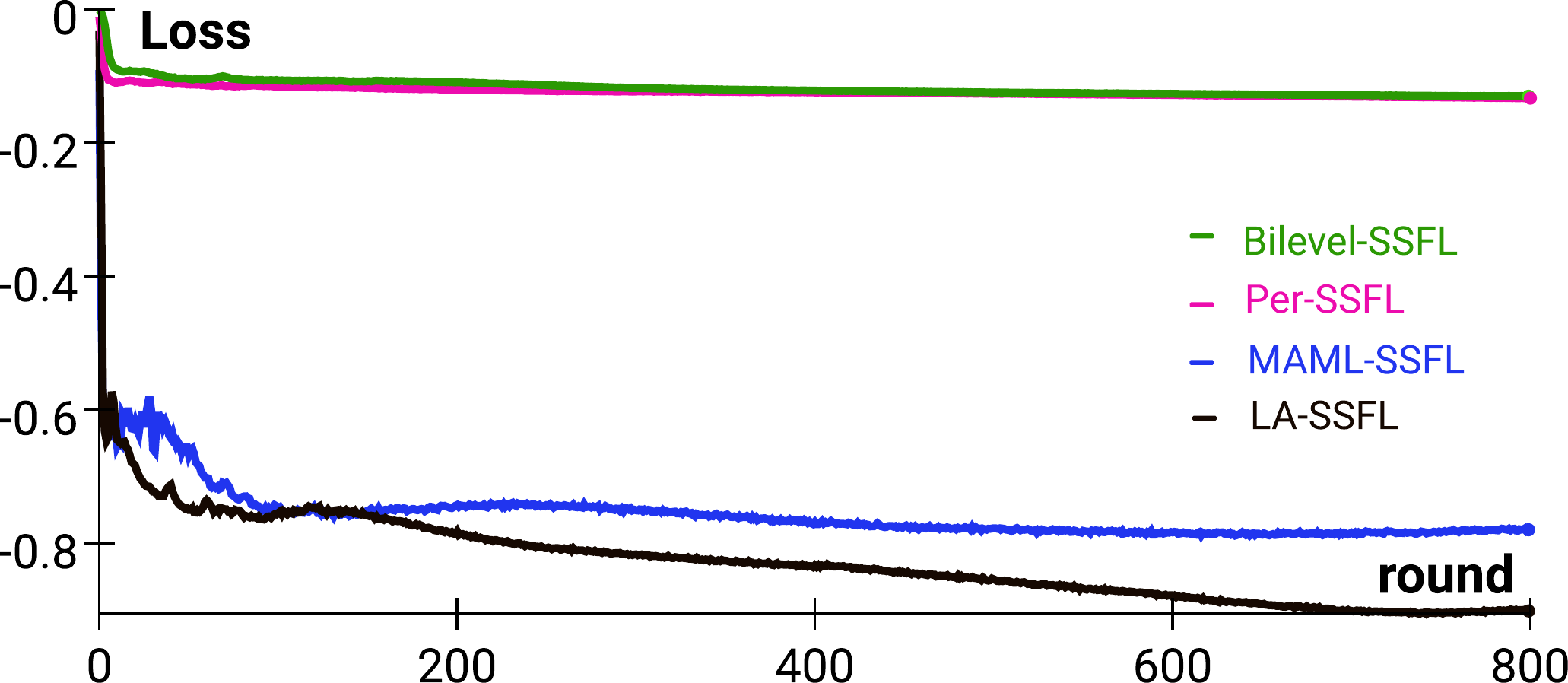}}}
\hspace{0.3cm}
\subfigure[\label{fig:1B} Averaged Personalized Accuracy]
{{\includegraphics[width=0.45\textwidth]{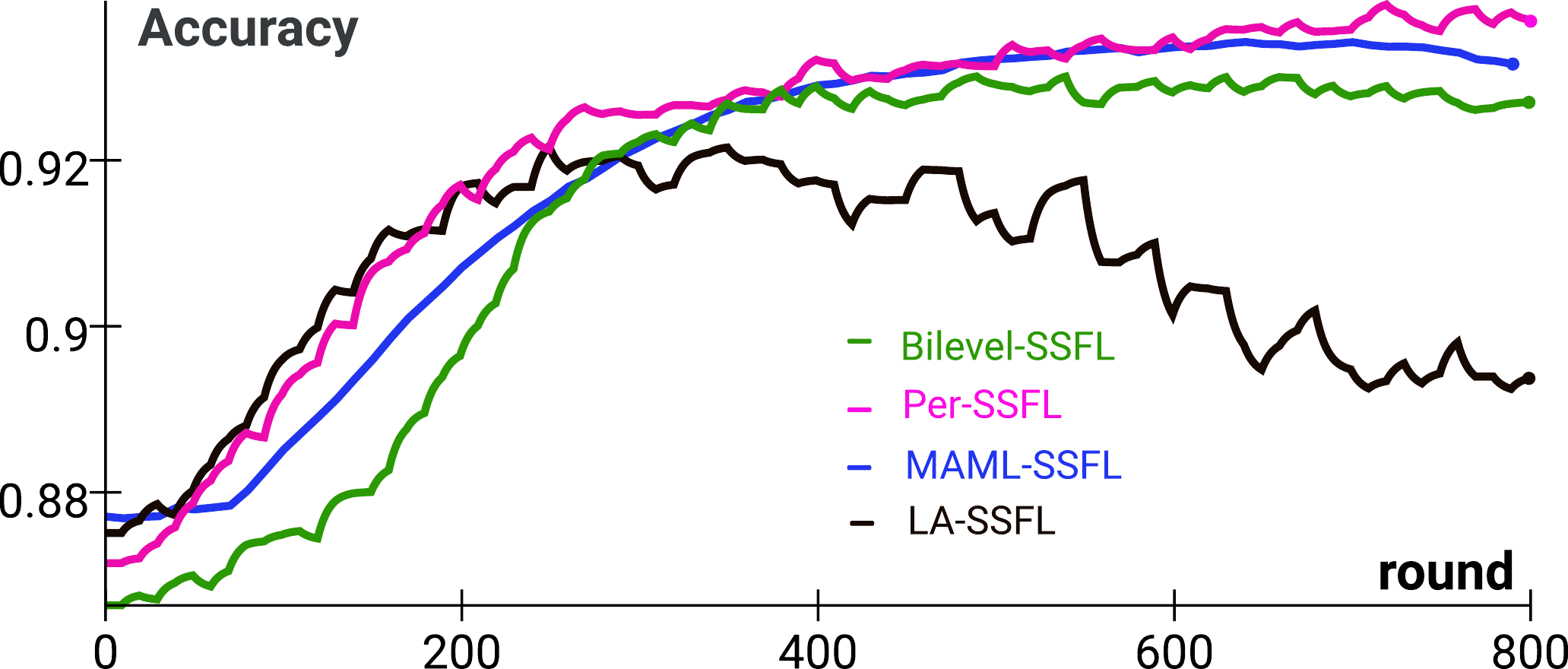}}}
\setlength{\belowcaptionskip}{-0.45cm}
\caption{Training and Evaluation using SSFL}
\label{fig:SSFL_without_per}
\end{figure}

We illustrate our results in Figure \ref{fig:SSFL_without_per} and Table \ref{table:per-ssfl}. To confirm the convergence, we draw loss curves for all methods in Figure \ref{fig:SSFL_without_per}(b) (note that they have different scaled values due to the difference of their loss functions). Figure \ref{fig:SSFL_without_per}(b) indicates that \texttt{Per-SSFL} performs best among all methods. \texttt{MAML-SSFL} is also a suggested method since it obtains comparable accuracy. \texttt{LA-SSFL} is a practical method, but it does not perform well in the self-supervised setting. In Figure \ref{fig:SSFL_without_per}(b), the averaged personalized accuracy of \texttt{LA-SSFL} diverges in the latter phase. Based on \texttt{BiLevel-SSFL}'s result, we can conclude such a method is not a strong candidate for personalization, though it shares similar objective functions as \texttt{Per-SSFL}. This indicates that regularization through representations encoded by \texttt{SimSiam} outperforms regularization through weights.


\subsection{Performance Analysis}

\subsubsection{Role of Batch Size}
\label{sec:batch_size}
\begin{wrapfigure}{r}{0.35\textwidth}
  \begin{center}
    \includegraphics[width=0.35\textwidth]{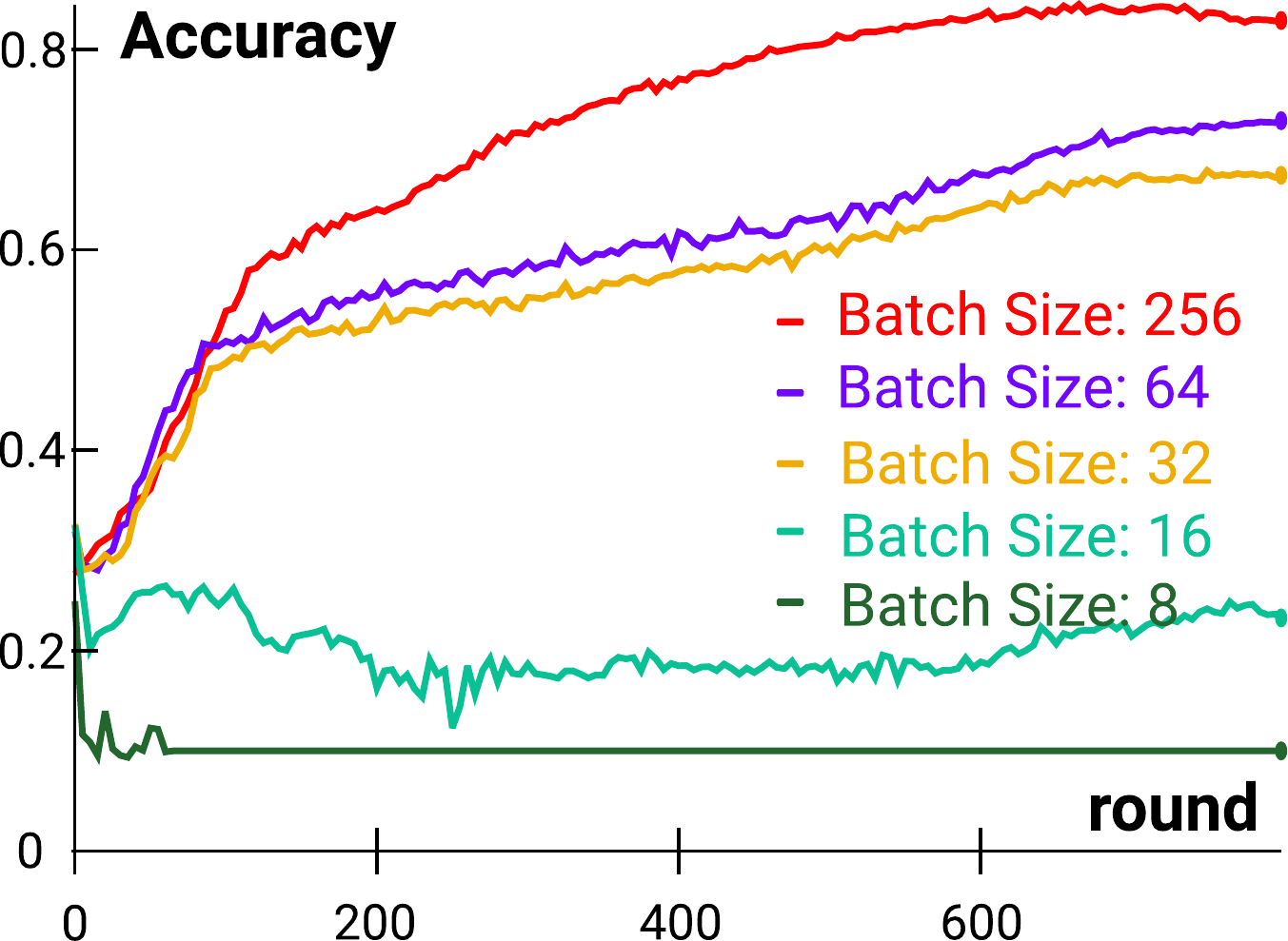}
  \end{center}
  \caption{Results for batch sizes}
  \label{fig:batch_size1}
\end{wrapfigure}
FL typically requires a small batch size to enable practical training on resource-constrained edge devices. Therefore, understanding the role of batch size in \texttt{SSFL} is essential to practical deployment and applicability. To investigate this, we use different batch sizes and tune the learning rate to find the optimal accuracy for each one. The results in Figure \ref{fig:batch_size1} show that SSFL requires a large batch size (256); otherwise, it reduces the accuracy or diverges during training. 
Since system efficiency is not the focus of this paper, we use gradient accumulation, which is a simple yet effective method. We fix the batch size at 32 and use accumulation step 8 for all experiments. For an even larger batch size (e.g., 512), the memory cost is significant, though there is no notable gain in accuracy. Therefore, we discontinue the search for batch sizes larger than larger than 256. A more advanced method includes batch-size-one training and knowledge distillation. We defer the discussion to Appendix \ref{app:discussion}.

\subsubsection{On Different Degrees of Non-I.I.D.ness} 
\label{exp:non-IID}

We investigate the impact of the degree of data heterogeneity on the SSFL performance.
We compare the performance between $\alpha=0.1$ and $\alpha=0.5$.
These two settings provide a non-negligible gap in the label distribution in each client (see our visualization in Appendix \ref{app:data}).
Figure \ref{fig:non-IID} and \ref{fig:non-IID Loss} shows the learning curve comparisons.
It is clearly observed that the higher degree of data heterogeneity makes it converge more slowly, adversely affecting the accuracy.

\begin{figure*}[h!]
  \centering
  \begin{minipage}{.65\textwidth}
    \centering
    \subfigure[\label{fig:non-IID} Averaged Personalized Accuracy]
      {\includegraphics[width=0.5\textwidth]{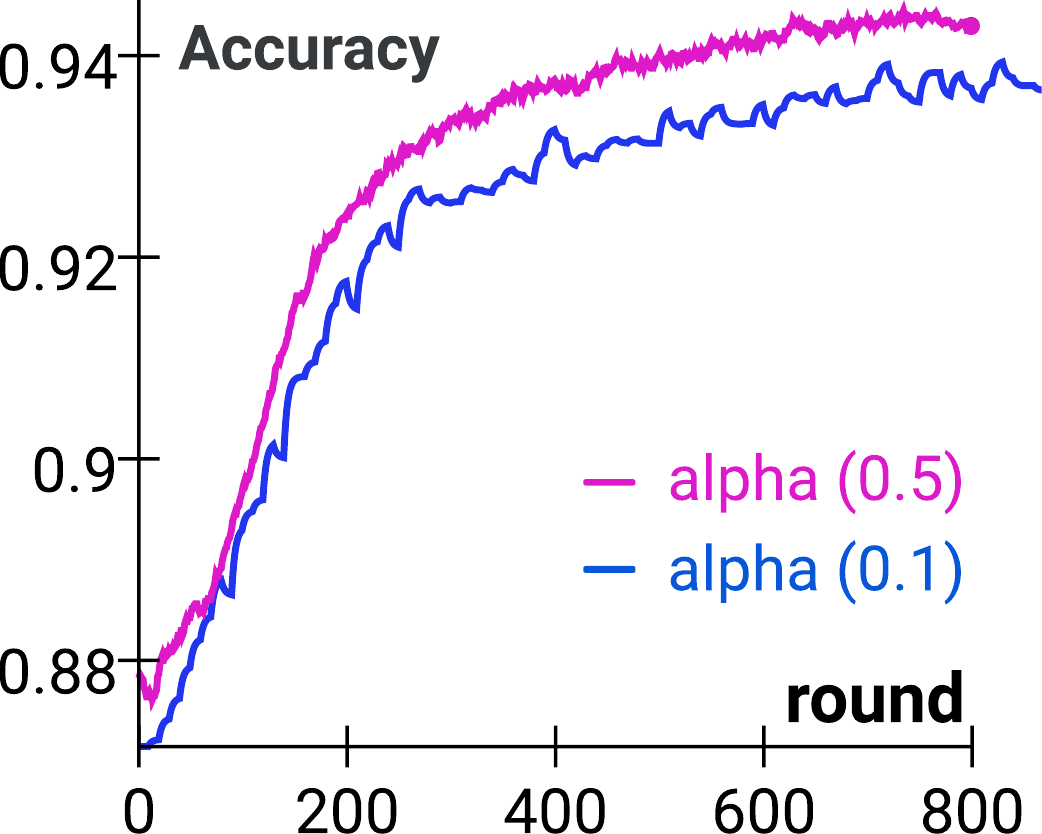}}
      \hspace{0.2cm}
      \subfigure[\label{fig:non-IID Loss} Training Loss]
      {\includegraphics[width=0.45\textwidth]{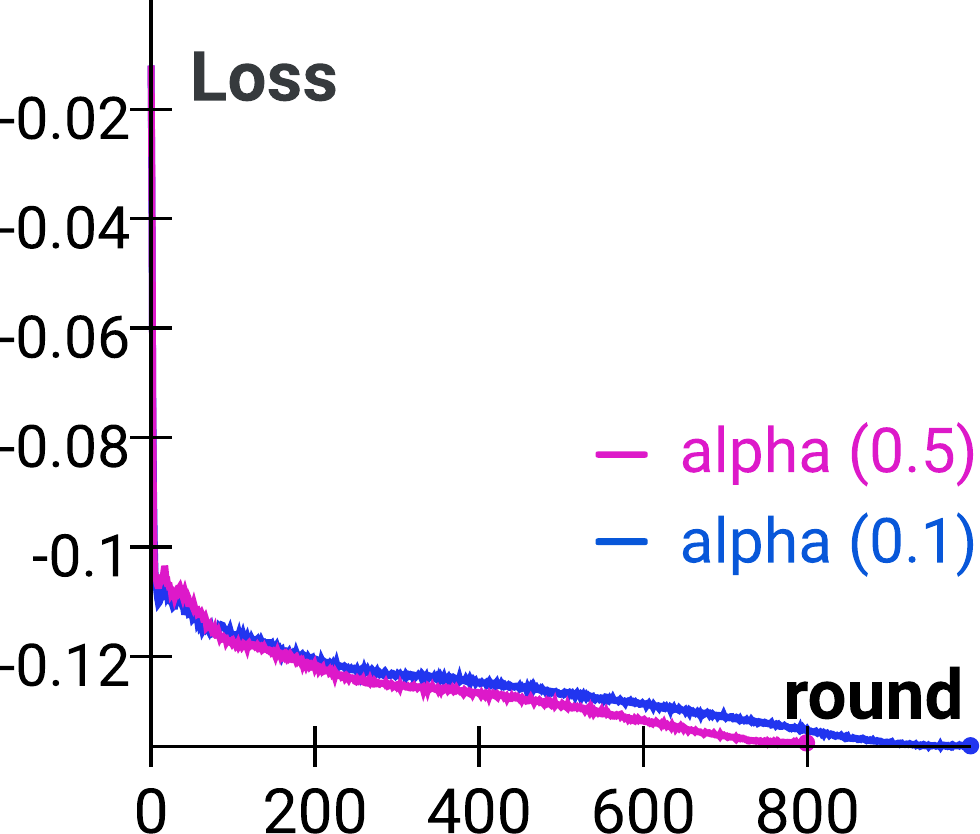}}
    \captionof{figure}{Evaluation on Different Degress of Non-I.I.D.ness}
  \end{minipage}
  \hspace{0.2cm}
  \begin{minipage}{.3\textwidth}
    \centering
      {\includegraphics[width=\columnwidth]{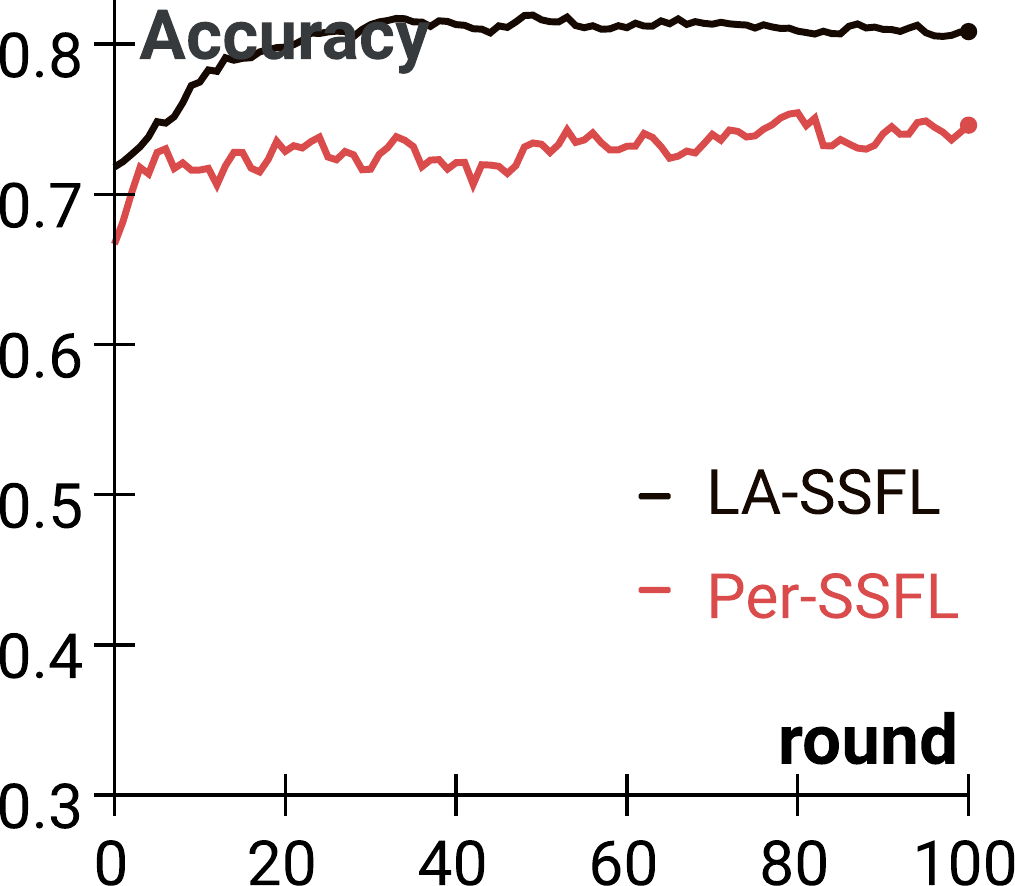}}
    \captionof{figure}{Understanding the Evaluation Protocol}
\label{exp:eval_protocol}
  \end{minipage}
\end{figure*}

\subsubsection{Understanding the Linear Evaluation of Personalized Encoders}
\label{sec:exp-eval-protocol}

As we discussed in \ref{sec:protocol}, in \texttt{SSFL}, we can easily verify the quality of the \texttt{SimSiam} encoder using federated linear evaluation; however, in \texttt{Per-SSFL}, each client learns a personalized \texttt{SimSiam} encoder. Such heterogeneity in diverse encoders makes a fair evaluation difficult. To demonstrate this, we run experiments with naive federated linear evaluation on personalized encoders and surprisingly find that such an evaluation protocol downgrades the performance. As shown in Figure \ref{exp:eval_protocol}, the federated linear evaluation for \texttt{Per-SSFL} performs worse than even \texttt{LA-SSFL}. This may be attributed to the fact that the naive aggregation drags close to the parameter space of all heterogeneous encoders, making the encoder degenerate in terms of personalization. 

\section{Related Works}
\textbf{Federated Learning (FL) with Personalization.} 
pFedMe \citep{dinh_personalized_2020}, perFedAvg \citep{fallah2020personalized}, and Ditto \citep{li_ditto_2021} are some representative works in this direction. However, these methods all have a strong assumption that users can provide reliable annotations for their private and sensitive data, which we argue to be very unrealistic and impractical. 
\\ \textbf{Label deficiency in FL.} There are a few related works to tackle label deficiency in FL \citep{liu2020rc,long2020fedsemi,itahara2020distillation,jeong2020federated,liang2021self,zhao2020semi,zhang2020improving}. Compared to these works, our proposed SSFL does not use any labels during training. FedMatch \citep{jeong2020federated} and \texttt{FedCA} \citep{zhang2020federated} requires additional communication costs to synchronize helper models or public labeled dataset.
\citep{saeed2020federated} addresses the fully unsupervised challenge on small-scale sensor data in IoT devices. However, compared to our work, it uses the Siamese networks proposed around thirty years ago \citep{bromley1993signature}, lacking consideration on the advance in the past two years (i.e., \texttt{SimCLR} \citep{chen2020simple}, \texttt{SwAV}\citep{caron2021unsupervised}, \texttt{BYOL} \citep{grill2020bootstrap}, and \texttt{SimSiam} \citep{chen2020exploring}). Moreover, these works does not have any design for learning personalized models.


\section{Conclusion}
We propose Self-supervised Federated Learning (SSFL) framework and a series of algorithms under this framework towards addressing two challenges: data heterogeneity and label deficiency. \texttt{SSFL} can work for both global model training and personalized model training. We conduct experiments on a synthetic non-I.I.D. dataset based on CIFAR-10 and the intrinsically non-I.I.D. GLD-23K dataset. Our experimental results demonstrate that \texttt{SSFL} can work reliably and achieves reasonable evaluation accuracy that is suitable for use in various applications.

\clearpage
\bibliography{iclr2022_conference}
\bibliographystyle{unsrtnat}

\clearpage
\appendix
\section*{Appendix}


\section{Comparison of Self-supervised Learning Frameworks}
\label{app:ssl}

We compare state-of-the-art self-supervised learning frameworks (\texttt{SimCLR}, \texttt{SwAV}, \texttt{BYOL}) with \texttt{SimSiam} \citep{chen2020exploring} in light of federated learning.

We choose \texttt{SimSiam} \citep{chen2020exploring} because it requires a much smaller batch to perform normally. In the centralized setting, for each method to reach an accuracy level similar to that of \texttt{SimSiam}, a much larger batch size is necessary. Table \ref{tab:ssl-sota} adopted from \citep{chen2020exploring} provides a brief comparison between all listed self-supervised learning frameworks.

\begin{table}[h!]
\centering
\small
\begin{tabular}{l c c c | c c c c}
method
& {\begin{tabular}{c} {batch} \\ {size} \end{tabular}}
& {\begin{tabular}{c} {negative} \\ {pairs} \end{tabular}}
& {\begin{tabular}{c} {momentum} \\ {encoder} \end{tabular}}
& 100 ep & 200 ep & 400 ep & 800 ep \\
\hline
SimCLR (repro.+) & 4096 & \checkmark & &
66.5 & 68.3 & 69.8 & 70.4 \\
BYOL (repro.) & 4096 & & \checkmark &
66.5 & \textbf{70.6} & \textbf{73.2} & \textbf{74.3} \\
SwAV (repro.+) & 4096 & & &
66.5 & 69.1 & 70.7 & 71.8 \\
SimSiam & \textbf{256} & & & 
\textbf{68.1} & 70.0 & 70.8 & 71.3 \\
\end{tabular}
\caption{
\citep{chen2020exploring} \textbf{Comparisons on ImageNet linear classification}. All are based on \textbf{ResNet-50} pre-trained with \textbf{two 224$\times$224 views} in a centralized setting. Evaluation is on a single crop.  ``repro.'' denotes reproduction conducted by authors of SimSiam \citep{chen2020exploring}, and ``+'' denotes \emph{improved} reproduction v.s. original papers.}
\label{tab:ssl-sota}
\end{table}

\begin{figure}[h!]
\begin{center}
\includegraphics[width=.7\linewidth]{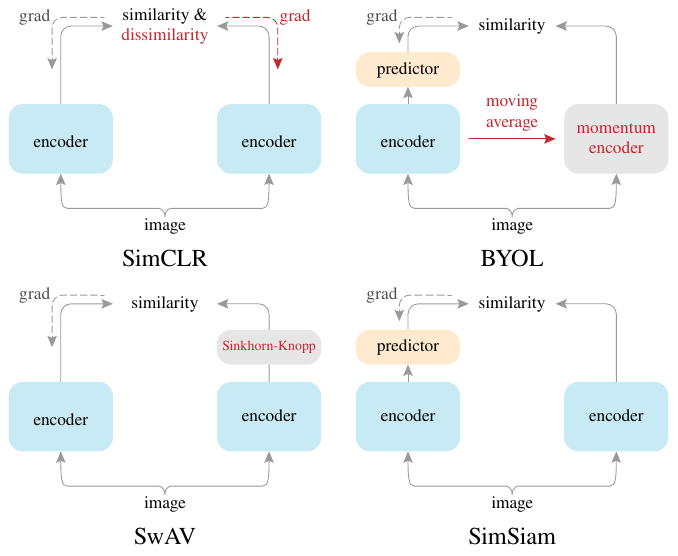}
\end{center}
\caption{
\citep{chen2020exploring} \textbf{Comparison on Siamese architectures}.
The encoder includes all layers that can be shared between both branches. 
The dashed lines indicate the gradient propagation flow.
In \texttt{BYOL}, \texttt{SwAV}, and \texttt{SimSiam}, the lack of a dashed line implies stop-gradient, and their symmetrization is not illustrated for simplicity. The components in red are those missing in SimSiam.
\label{fig:methodology}
}
\end{figure}

Another reason we prefer \texttt{SimSiam} \citep{chen2020exploring} as the basic framework to build \texttt{SSFL} is that the design of \texttt{SimSiam} simplifies all other baselines and also obtains a relatively higher accuracy. Figure \ref{fig:methodology} abstracts these methods. The ``encoder'' contains all layers that can be shared between both branches (e.g., backbone, projection MLP \citep{chen2020simple}, prototypes \citep{caron2021unsupervised}). The components in red are those missing in \texttt{SimSiam}.

\paragraph{SimCLR \citep{chen2020simple}.} 
\texttt{SimCLR} relies on negative samples (``dissimilarity'') to prevent collapsing.
\texttt{SimSiam} can be thought of as ``\texttt{SimCLR} without negatives". In every mini-batch, for any image, one augmented view of the same image is considered to be its positive sample, and the remaining augmented views of different images are considered to be its negative samples. A contrastive loss term is calculated to push positive samples together and negative samples away.

\paragraph{SwAV\citep{caron2021unsupervised}.} \texttt{SimSiam} is conceptually analogous to ``\texttt{SwAV} without online clustering".
SimSiam encourages the features of the two augmented views of the same image to be similar, while \texttt{SwAV} encourages features of the two augmented views of the same image to belong to the same cluster. An additional Sinkhorn-Knopp (SK) transform \citep{cuturi2013sinkhorn} is required for online clustering of \texttt{SwAV}. 
The authors of \texttt{SimSiam} \citep{chen2020exploring} build up the connection between \texttt{SimSiam} and \texttt{SwAV} by recasting a few components in \texttt{SwAV}.
(i) The shared prototype layer in \texttt{SwAV} can be absorbed into the Siamese encoder.
(ii) The prototypes were weight-normalized outside of gradient propagation in \citep{caron2021unsupervised};
the authors of \texttt{SimSiam} instead implement by full gradient computation \citep{DBLP:journals/corr/SalimansK16}.
(iii) The similarity function in \texttt{SwAV} is cross-entropy.
With these abstractions, a highly simplified \texttt{SwAV} illustration is shown in Figure~\ref{fig:methodology}.

\paragraph{BYOL \citep{grill2020bootstrap}.} \texttt{SimSiam} can be thought of as ``\texttt{BYOL} without the momentum encoder", subject to many implementation differences. Briefly, in \texttt{BYOL}, one head of the Siamese architecture used in \texttt{SimSiam} is replaced by the exponential moving average of the encoder. As the momentum encoder has an identical architecture to that of the encoder, the introduction of an additional momentum encoder doubles the memory cost of the model.

SSL's recent success is the inductive bias that ensures a good representation encoder remains consistent under different perturbations of the input (i.e. consistency regularization). The perturbations can be either domain-specific data augmentation (e.g. random flipping in the image domain) \citep{berthelot_mixmatch_2019, laine_temporal_2016, sajjadi_regularization_2016,berthelot_remixmatch_2020, hu_simple_2021}, drop out \citep{sajjadi_regularization_2016}, random max pooling \citep{sajjadi_regularization_2016}, or an adversarial transformation \citep{miyato_virtual_2019}. With this idea, a consistency loss $\mathcal{L}$ is defined to measure the quality of the representations without any annotations. 

\newpage
\section{Formulation and Pseudo Code for Algorithms Under SSFL Framework}
\label{app:algorithms}

Inspired by recent advances in personalized FL and self-supervised learning, we innovate several representative algorithms under \texttt{SSFL} framework. For each algorithm, we present its mathematical formulation and its pseudo code. 

\subsection{Per-SSFL}
For \texttt{Per-SSFL}, as the formulation and algorithm have already been presented in Equation \ref{eq:perssfl1} and Algorithm \ref{alg:pfss}, we provide a PyTorch style pseudo code in Algorithm \ref{alg:pfss-pytorch} for additional clarity.

\begin{algorithm}[h!]
\begin{lstlisting}[language=Python]
# F: global encoder
# H: global predictor
# f: local encoder
# h: local predictor

for x in loader:  # load a mini-batch x with n samples
    x1, x2 = aug(x), aug(x)  # random augmentation
    Z1, Z2 = F(x1), F(x2)  # global projections, n-by-d
    P1, P2 = H(Z1), H(Z2)  # global predictions, n-by-d
    
    L = D(P1, Z2) / 2 + D(P2, Z1) / 2  # global loss
    
    L.backward()  # back-propagate
    update(F, H)  # SGD update global model
    
    z1, z2 = f(x1), f(x2)  # local projections, n-by-d
    p1, p2 = h(z1), h(z2)  # local predictions, n-by-d
    
    l = D(p1, z2) / 2 + D(p2, z1) / 2  # local loss
    
    # distance between local and global representations
    l = l + λ * (D(p1, P1) + D(p1, P2) + D(p2, P1) + D(p2, P2)) / 4
    
    l.backward()  # back-propagate
    update(f, h)  # SGD update local model

def D(p, z):  # negative cosine similarity
    z = z.detach()  # stop gradient
    
    p = normalize(p, dim=1)  # l2-normalize
    z = normalize(z, dim=1)  # l2-normalize
    return -(p * z).sum(dim=1).mean()
\end{lstlisting}
    \caption{Per-SSFL PyTorch Style Pseudo Code}
    \label{alg:pfss-pytorch}
\end{algorithm}

\newpage
\subsection{Personalized SSFL with Local Adaptation (FedAvg-LA)}
\texttt{FedAvg-LA} apply \texttt{FedAvg} \citep{brendan2016} on the \texttt{SimSiam} loss $\mathcal{L}_\mathrm{SS}$ for each client to obtain a global model. We perform one step of SGD on the clients' local data for local adaption. The objective is defined in Equation \ref{eq:fedavg-la}, and the algorithm is provided in Algorithm \ref{alg:fedavg-la}.

\begin{equation}
\min_{\Theta,\mathcal{H}}\sum_{i=1}^{n}\frac{|D_{k}|}{|D|}\E_{\substack{\Tau\\
x\sim X_{i}
}
}\left[\left\Vert f_{\Theta}(\Tau(x))-\mathcal{H}_{x}\right\Vert _{2}^{2}\right]
\label{eq:fedavg-la}
\end{equation}

\input{listing/FedAvg-LA}

\newpage
\subsection{Personalized SSFL with MAML-SSFL}
\texttt{MAML-SSFL} is inspired by perFedAvg \citep{fallah_personalized_2020-2} and views the personalization on each devices as the inner loop of MAML \citep{finn_model-agnostic_2017}. It aims to learn an encoder that can be easily adapted to the clients' local distribution. During inference, we perform one step of SGD on the global model for personalization. The objective is defined in Equation \ref{eq:maml-ssfl}, and the algorithm is provided in Algorithm \ref{alg:maml-ssfl}.

\begin{equation}
    \begin{aligned}
        \min_{\Theta,\mathcal{H}}&\quad\sum_{i=1}^{n}\frac{|D_{k}|}{|D|}\E_{\substack{\Tau\\
        x\sim X_{i}
        }
        }\left[\left\Vert f_{\Theta^{\prime}}(\Tau(x))-\mathcal{H}_{x}\right\Vert _{2}^{2}\right]\\\text{ s.t. }&\quad\Theta^{\prime}=\Theta-\nabla_{\Theta}\sum_{i=1}^{n}\frac{|D_{k}|}{|D|}\E_{\substack{\Tau\\
        x\sim X_{i}
        }
        }\left[\left\Vert f_{\Theta}(\Tau(x))-\mathcal{H}_{x}\right\Vert _{2}^{2}\right]
    \end{aligned}
    \label{eq:maml-ssfl}
\end{equation}

\input{listing/MAML-SSFL}

\newpage
\subsection{Personalized SSFL with BiLevel-SSFL}
Inspired by Ditto \citep{li_ditto_2021}, \texttt{BiLevel-SSFL} learns personalized encoders on each client by restricting the parameters of all personalized encoders to be close to a global encoder independently learned by weighted aggregation. The objective is defined in Equation \ref{eq:bilevel-ssfl}, and the algorithm is provided in Algorithm \ref{alg:bilevel-ssfl}.

\begin{equation}
    \begin{aligned}
        \min_{\theta_{k},\eta_{k}}&\quad\E_{\substack{\Tau\\
        x\sim X_{k}
        }
        }\left[\left\Vert f_{\theta_{k}}(\Tau(x))-\eta_{k,x}\right\Vert _{2}^{2}+\frac{\lambda}{2}\left\Vert \theta_{k}-\Theta_{x}^{*}\right\Vert _{2}^{2}\right]\\\text{s.t.}&\quad\Theta^{*},\mathcal{H}^{*}\in\arg\min_{\Theta,\mathcal{H}}\sum_{i=1}^{n}\frac{|D_{k}|}{|D|}\E_{\substack{\Tau\\
        x\sim X_{i}
        }
        }\left[\left\Vert f_{\Theta}(\Tau(x))-\mathcal{H}_{x}\right\Vert _{2}^{2}\right]
    \end{aligned}
    \label{eq:bilevel-ssfl}
\end{equation}

\input{listing/BeLevel-SSFL}

\newpage
\section{Distributed Training System for SSFL}
\label{app:system}

\begin{figure}[h!]
\centering
{{\includegraphics[width=1\textwidth]{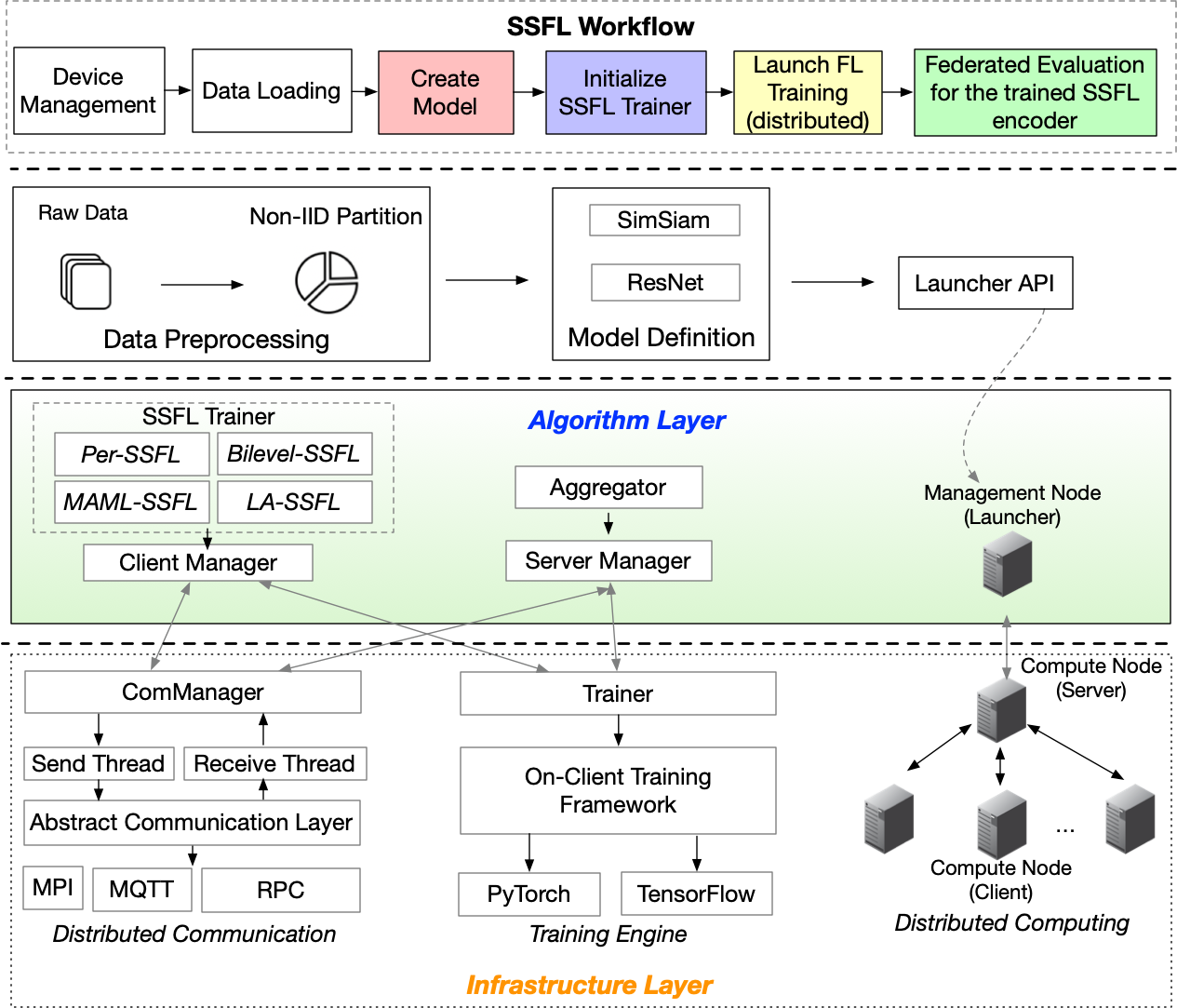}}}
\hspace{0.5cm}
\caption{Distributed Training System for \texttt{SSFL} framework }
\label{fig:system_design}
\end{figure}
We develop a distributed training system for our \texttt{SSFL} framework which contains three layers. In the infrastructure layer, communication backends such as MPI are supported to facilitate the distributed computing. We abstract the communication as \texttt{ComManager} to simplify the message passing between the client and the server. \texttt{Trainer} reuses APIs from PyTorch to handle the model optimizations such as forward propagation, loss function, and back propagation. In the algorithm layer, \texttt{Client Manager} and \texttt{Server Manager} are the entry points of the client and the server, respectively. The client managers incorporates various \texttt{SSFL} trainers, including \texttt{Per-SSFL}, \texttt{MAML-SSFL}, \texttt{BiLevel-SSFL}, and \texttt{LA-SSFL}. The server handles the model aggregation using \texttt{Aggregator}. We design simplified APIs for all of these modules. With the abstraction of the infrastructure and algorithm layers, developers can begin FL training by developing a workflow script that integrates all modules (as the ``SSFL workflow'' block shown in the figure). Overall, we found that this distributed training system accelerates our research by supporting parallel training, larger batch sizes, and easy-to-customize APIs, which cannot be achieved by a simple single-process simulation.

\newpage
\section{Experimental Results on GLD-23K Dataset}
\label{app:gld23k}
We also evaluate the performance of SSFL on GLD-23K dataset. We use 30\% of the original local training dataset as the local test dataset and filter out those clients that have a number of samples less than 100. Due to the natural non-I.I.D.ness of GLD-23K dataset, we only evaluate the Per-SSFL framework. The results are summarized in Table \ref{table:per-ssfl-gld23k}. \textit{Note: we plan to further explore more datasets and run more experiments; thus we may report more results during the rebuttal phase.}

\begin{table}[h!]
\centering

\caption{Evaluation Accuracy for Various Per-SSFL Methods.}
\label{table:per-ssfl-gld23k}
\resizebox{0.5\textwidth}{!}{
\begin{threeparttable}
\begin{tabular}{ccc}
\toprule
Method & KNN Indicator & Evaluation \\
\midrule
\texttt{LA-SSFL} & 0.6011 & 0.4112\\
\texttt{MAML-SSFL} & 0.6237 & 0.4365\\
\texttt{BiLevel-SSFL} & 0.6195 & 0.4233\\
\texttt{Per-SSFL} & 0.6371 & 0.4467\\
\bottomrule    
\end{tabular}
\begin{tablenotes}[para,flushleft]
      \footnotesize
      \item *Note: the accuracy on supervised federated training using FedAvg is around 47\%
    \end{tablenotes}
\end{threeparttable}}
\setlength{\belowcaptionskip}{-0.2cm}
\end{table}


\section{Extra Experimental Results and Details}

\subsection{Visualization of Non-I.I.D. dataset}
\label{app:data}
\begin{figure}[h!]
\centering
\subfigure[\label{fig:1A} Sample Number Distribution]
{{\includegraphics[width=0.45\textwidth]{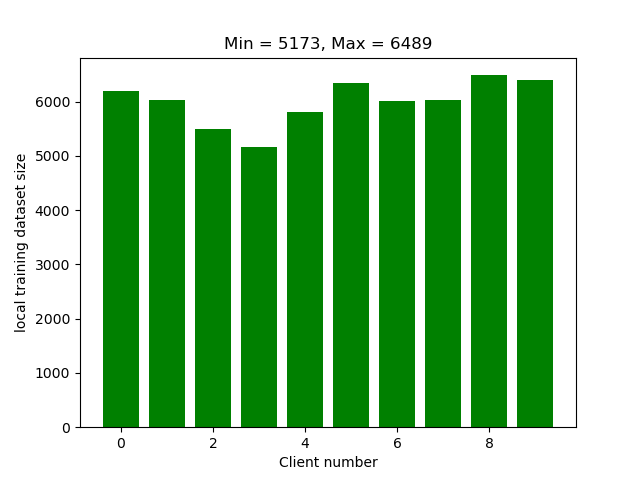}}}
\hspace{0.5cm}
\subfigure[\label{fig:1B} Label Distribution (deeper color stands for more samples]
{{\includegraphics[width=.45\textwidth]{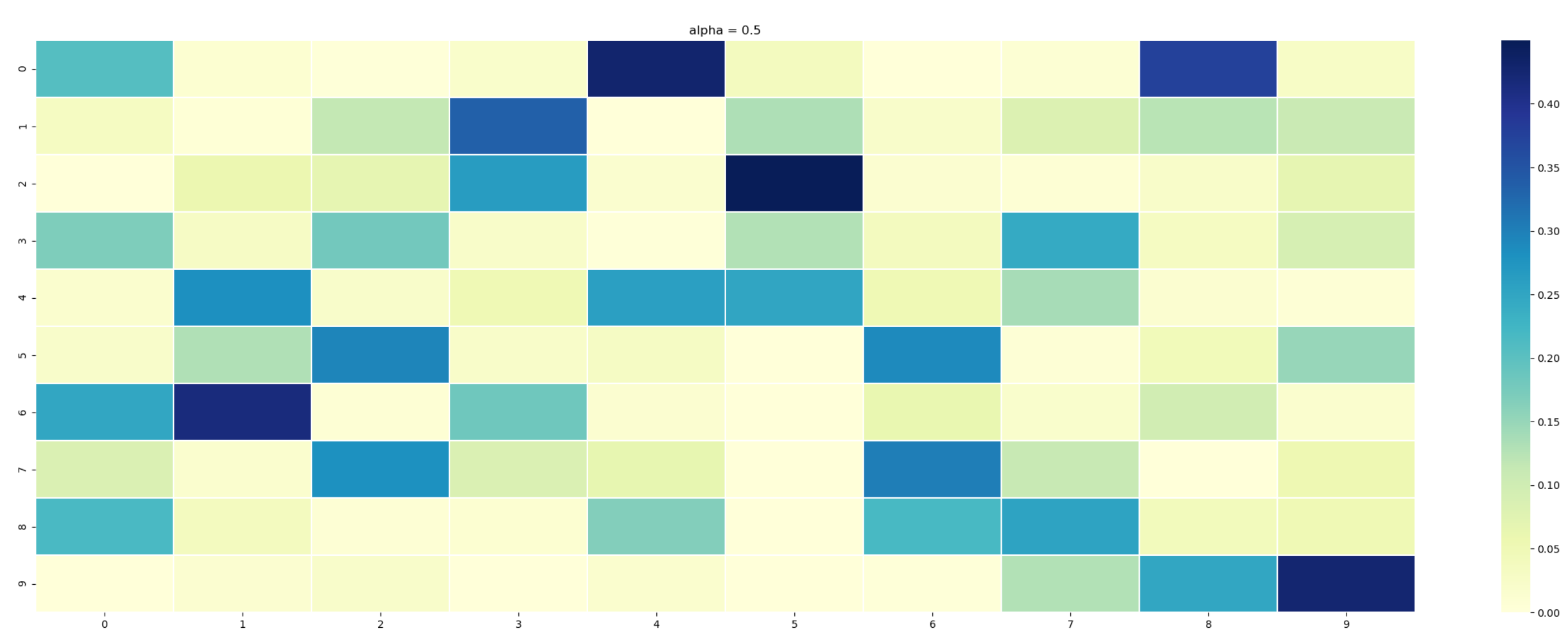}}}
\caption{Visualization for non-I.I.D. synthesized using CIFAR-10}
\label{fig:SSFL_without_per}
\end{figure}

\begin{figure}[h!]
\centering
\subfigure[\label{fig:1A}  Sample Number Distribution (X-axis: Client Index; Y-axis: Number of Training Samples)]
{{\includegraphics[width=0.45\textwidth]{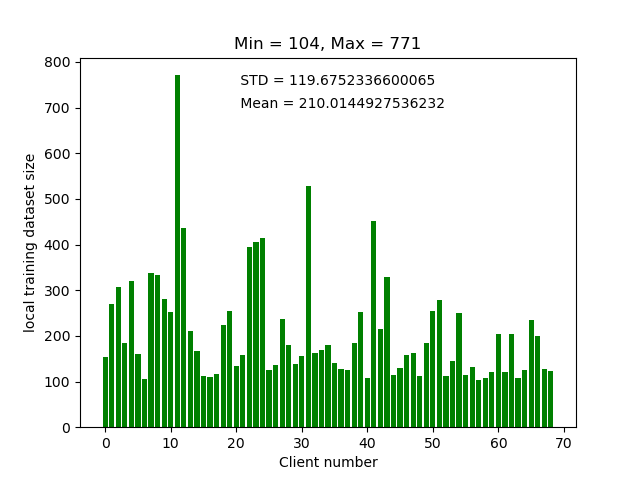}}}
\hspace{0.5cm}
\subfigure[\label{fig:1B} Sample Number Distribution (X-axis: Number of Training Samples; Y-axis: Number of Clients)]
{{\includegraphics[width=0.45\textwidth]{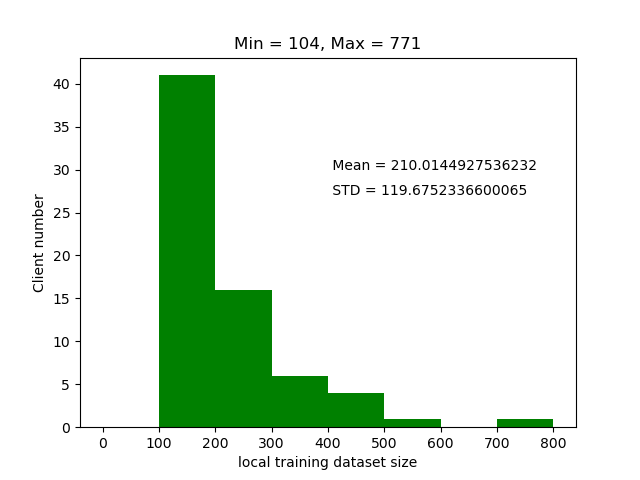}}}
\caption{Visualization for non-I.I.D. on GLD-23K}
\label{fig:SSFL_without_per}
\end{figure}
\subsection{Hyper-parameters}
\label{app:hps}

\begin{table}[h!]
    \centering
    \caption{Hyper-parameters for Section \ref{sec:exp-ssfl}}
    \label{table:para-data}
    \resizebox{0.7\textwidth}{!}{
   \begin{tabular}{lcc}
    \toprule
       Method & Learning Rate & Local Optimizer \\
        \midrule
        SSFL (I.I.D)& 0.1  & SGD with Momemtum (0.9) \\
        SSFL (non-I.I.D)& 0.1 & SGD with Momemtum (0.9) \\
       \bottomrule
       \end{tabular}}
 \end{table}
  
  \begin{table}[h!]
    \centering
    \label{table:para-alpha}
 \caption{Hyper-parameters for Section \ref{exp:non-IID}}
    \resizebox{0.7\textwidth}{!}{
   \begin{tabular}{lccc}
    \toprule
       Method & Learning Rate & $\lambda$ & Local Optimizer \\
        \midrule
        Per-SSFL ($\alpha=0.1$)& 0.03 &0.1 & SGD with Momemtum (0.9)\\
        Per-SSFL ($\alpha=0.5$)& 0.03 &0.1 & SGD with Momemtum (0.9)\\
       \bottomrule
       \end{tabular}}
  \end{table}

  \begin{table}[h!]
    \centering
     \caption{Hyper-parameters for experimental results in Section \ref{sec:exp-per-4-2}}
      \label{table:para-method}
    \resizebox{.7\textwidth}{!}{
  \begin{tabular}{lccc}
\toprule
Method & Learning Rate & $\lambda$ & Local Optimizer \\
\midrule
LA-SSFL & 0.1 & 1 & SGD with Momemtum (0.9) \\
MAML-SSFL & 0.03 & 1& SGD with Momemtum (0.9) \\
BiLevel-SSFL & 0.1 & 1& SGD with Momemtum (0.9) \\
Per-SSFL& 0.03 & 0.1& SGD with Momemtum (0.9) \\
\bottomrule    
    \end{tabular}}
    \end{table}
All experiments set the local epoch number as 1, round number as 800, batch size as 256 (batch size 32 with 8 gradient accumulation steps).

\section{Discussion}
\label{app:discussion}
To overcome the large batch size requirement in \texttt{SSFL} and practical FL edge training, one direction is to use efficient DNN models such as EfficientNet \citep{tan2019efficientnet} and MobileNet \citep{howard2017mobilenets} as the backbone of SimSiam. However, we tested its performance under our framework and found that the performance downgrades to a level of accuracy that is not useful (less than 60\%). A recent work in centralized self-supervised learning mitigates these models' accuracy gap by knowledge distillation, which works in a centralized setting but is still not friendly to FL since KD requires additional resources for the teacher model. In practice, we can also explore batch size 1 training \citep{cai2020tinytl} at the edge, which dramatically reduces the memory cost with additional training time.
\end{document}

%% file: math_commands.tex
\usepackage{amsmath,amsfonts,bm}

\def\eqref#1{equation~\ref{#1}}

\def\1{\bm{1}}

\DeclareMathAlphabet{\mathsfit}{\encodingdefault}{\sfdefault}{m}{sl}
\SetMathAlphabet{\mathsfit}{bold}{\encodingdefault}{\sfdefault}{bx}{n}

\newcommand{\E}{\mathbb{E}}



%% file: listing/PerFedSimSiam.tex
\begin{center}
\scalebox{0.85}{
\begin{minipage}{\linewidth}

\begin{algorithm}[H]
\small
\SetKwInOut{Input}{input}\SetKwInOut{Output}{return}
\Input{$K, T, \lambda, \Theta^{(0)}, \{\theta_i^{(0)}\}_{k\in[K]}, s\text{: number of local iteration}, \beta\text{: learning rate}$}
\For{$t=0,\ldots,T-1$}{
    Server randomly selects a subset of devices $S^{(t)}$\\
    Server sends the current global model $\Theta^{(t)}$ to $S^{(t)}$\\
    \For{device $k \in S^{(t)}$ in parallel}{
        \begin{tikzpicture} [remember picture, overlay]
            \node at (3pt, 2pt) (a) {};
            \node at (352pt,-87.5pt) (b) {};
            \node[line width=0pt, rounded corners=2pt, fill=blue!18, fill opacity=0.3, rectangle, fit=(a) (b)] (x) {};
            \node[anchor=south east,inner sep=1pt] at (x.south east) {\scriptsize$\clientOpt$};
        \end{tikzpicture}%
        Sample mini-batch $B_k$ from local dataset $D_k$, and do $s$ local iterations\\
        \tcc{Optimize the global parameter $\Theta$ locally}
        $Z_1, Z_2 \leftarrow f_{\Theta^{(t)}}(\Tau(B_k)), f_{\Theta^{(t)}}(\Tau(B_k))$\\
        $P_1, P_2 \leftarrow h_{\Theta^{(t)}}(Z_1), h_{\Theta^{(t)}}(Z_2)$\\
        $\Theta_k^{(t)} \leftarrow \Theta^{(t)} - \beta \nabla_{\Theta^{(t)}} \frac{\D(P_1, \widehat{Z_2}) + \D(P_2, \widehat{Z_1})}{2}$, where $\widehat{\cdot}$ stands for stop-gradient\\
        \tcc{Optimize the local parameter $\theta_k$}
        $z_1, z_2 \leftarrow f_{\theta_k}(\Tau(B_k)), f_{\theta_k}(\Tau(B_k))$\\
        $p_1, p_2 \leftarrow h_{\theta_k}(z_1), h_{\theta_k}(z_2)$\\
        $\theta_k \leftarrow \theta_k - \beta \nabla_{\theta_k}\left( \frac{\D(p1, \widehat{z_2}) + \D(p2, \widehat{z_1})}{2} 
            + \lambda \frac{\D(p1, P_1) + \D(p1, P_2) + \D(p2, P_1) + \D(p2, P_2)}{4} \right)$\\
        Send $\Delta_k^{(t)}:=\Theta_k^{(t)}-\Theta^{(t)}$ back to server
    }
    \begin{tikzpicture} [remember picture, overlay]
        \node at (5pt, 5pt) (a) {};
        \node at (367pt, 0pt) (b) {};
        \node[line width=0pt, rounded corners=2pt, fill=green!18, fill opacity=0.3, rectangle, fit=(a) (b)] (x) {};
        \node[anchor=south east,inner sep=1pt] at (x.south east) {\scriptsize$\serverOpt$};
    \end{tikzpicture}%
    $\Theta^{(t+1)} \leftarrow \Theta^{(t)} + \sum_{k \in S^{(t)}} \frac{|D_k|}{|D|} \Delta_k^{(t)}$
}
\Output{$\{\theta_i\}_{i\in[n]}, \Theta^{(T)}$}
\caption{Per-SSFL}
\label{alg:pfss}
\end{algorithm}

\end{minipage}%
}
\end{center}

%% file: listing/FedAvg-LA.tex
\begin{algorithm}[h!]
\small
\SetKwInOut{Input}{input}\SetKwInOut{Output}{return}
\Input{$K, T, \lambda, \Theta^{(0)}, \{\theta_i^{(0)}\}_{k\in[K]}, s\text{: number of local iteration}, \beta\text{: learning rate}$}
\For{$t=0,\ldots,T-1$}{
    Server randomly selects a subset of devices $S^{(t)}$\\
    Server sends the current global model $\Theta^{(t)}$ to $S^{(t)}$\\
    \For{device $k \in S^{(t)}$ in parallel}{
        \begin{tikzpicture} [remember picture, overlay]
            \node at (3pt, 2pt) (a) {};
            \node at (352pt,-41.5pt) (b) {};
            \node[line width=0pt, rounded corners=2pt, fill=blue!18, fill opacity=0.3, rectangle, fit=(a) (b)] (x) {};
            \node[anchor=south east,inner sep=1pt] at (x.south east) {\scriptsize$\clientOpt$};
        \end{tikzpicture}%
        Sample mini-batch $B_k$ from local dataset $D_k$, and do $s$ local iterations\\
        \tcc{Optimize the global parameter $\Theta$ locally}
        $Z_1, Z_2 \leftarrow f_{\Theta^{(t)}}(\Tau(B_k)), f_{\Theta^{(t)}}(\Tau(B_k))$\\
        $P_1, P_2 \leftarrow h_{\Theta^{(t)}}(Z_1), h_{\Theta^{(t)}}(Z_2)$\\
        $\Theta_k^{(t)} \leftarrow \Theta^{(t)} - \beta \nabla_{\Theta^{(t)}} \frac{\D(P_1, \widehat{Z_2}) + \D(P_2, \widehat{Z_1})}{2}$, where $\widehat{\cdot}$ stands for stop-gradient\\
        Send $\Delta_k^{(t)}:=\Theta_k^{(t)}-\Theta^{(t)}$ back to server
    }
    \begin{tikzpicture} [remember picture, overlay]
        \node at (5pt, 5pt) (a) {};
        \node at (367pt, 0pt) (b) {};
        \node[line width=0pt, rounded corners=2pt, fill=green!18, fill opacity=0.3, rectangle, fit=(a) (b)] (x) {};
        \node[anchor=south east,inner sep=1pt] at (x.south east) {\scriptsize$\serverOpt$};
    \end{tikzpicture}%
    $\Theta^{(t+1)} \leftarrow \Theta^{(t)} + \sum_{k \in S^{(t)}} \frac{|D_k|}{|D|} \Delta_k^{(t)}$
}
\Output{$\{\theta_i\}_{i\in[n]}, \Theta^{(T)}$}
\caption{FedAvg-LA}
\label{alg:fedavg-la}
\end{algorithm}

%% file: listing/MAML-SSFL.tex
\begin{algorithm}[h!]
\small
\SetKwInOut{Input}{input}\SetKwInOut{Output}{return}
\Input{$K, T, \lambda, \Theta^{(0)}, \{\theta_i^{(0)}\}_{k\in[K]}, s\text{: number of local iteration}, \beta\text{: learning rate}, M$}
\For{$t=0,\ldots,T-1$}{
    Server randomly selects a subset of devices $S^{(t)}$\\
    Server sends the current global model $\Theta^{(t)}$ to $S^{(t)}$\\
    \For{device $k \in S^{(t)}$ in parallel}{
        \begin{tikzpicture} [remember picture, overlay]
            \node at (3pt, 2pt) (a) {};
            \node at (352pt,-126pt) (b) {};
            \node[line width=0pt, rounded corners=2pt, fill=blue!18, fill opacity=0.3, rectangle, fit=(a) (b)] (x) {};
            \node[anchor=south east,inner sep=1pt] at (x.south east) {\scriptsize$\clientOpt$};
        \end{tikzpicture}%
        Sample mini-batch $B_k, B_k^\prime$ from local dataset $D_k$, and do $s$ local iterations\\
        \tcc{Inner loop update}
        $\Theta_k^{\prime (t)} \leftarrow \Theta^{(t)}$\\
        \For{$m = 0, \ldots, M-1$}{
            $Z_1^\prime, Z_2^\prime \leftarrow f_{\Theta^{\prime (t)}}(\Tau(B_k^\prime)), f_{\Theta^{\prime (t)}}(\Tau(B_k^\prime))$\\
            $P_1^\prime, P_2^\prime \leftarrow h_{\Theta^{\prime (t)}}(Z_1^\prime), h_{\Theta^{\prime (t)}}(Z_2^\prime)$\\
            $\Theta_k^{\prime (t)} \leftarrow \Theta_k^{\prime (t)} - \beta \nabla_{\Theta_k^{\prime (t)}} \frac{\D(P_1^\prime, \widehat{Z_2^\prime}) + \D(P_2^\prime, \widehat{Z_1^\prime})}{2}$, where $\widehat{\cdot}$ stands for stop-gradient\\
        }
        \tcc{Outer loop update}
        $Z_1, Z_2 \leftarrow f_{\Theta^{\prime (t)}}(\Tau(B_k)), f_{\Theta^{\prime (t)}}(\Tau(B_k))$\\
        $P_1, P_2 \leftarrow h_{\Theta^{\prime (t)}}(Z_1), h_{\Theta^{\prime (t)}}(Z_2)$\\
        $\Theta_k^{(t)} \leftarrow \Theta^{(t)} - \beta \nabla_{\Theta^{(t)}} \frac{\D(P_1, \widehat{Z_2}) + \D(P_2, \widehat{Z_1})}{2}$\\
        Send $\Delta_k^{(t)}:=\Theta_k^{(t)}-\Theta^{(t)}$ back to server
    }
    \begin{tikzpicture} [remember picture, overlay]
        \node at (5pt, 5pt) (a) {};
        \node at (367pt, 0pt) (b) {};
        \node[line width=0pt, rounded corners=2pt, fill=green!18, fill opacity=0.3, rectangle, fit=(a) (b)] (x) {};
        \node[anchor=south east,inner sep=1pt] at (x.south east) {\scriptsize$\serverOpt$};
    \end{tikzpicture}%
    $\Theta^{(t+1)} \leftarrow \Theta^{(t)} + \sum_{k \in S^{(t)}} \frac{|D_k|}{|D|} \Delta_k^{(t)}$
}
\Output{$\{\theta_i\}_{i\in[n]}, \Theta^{(T)}$}
\caption{MAML-SSFL}
\label{alg:maml-ssfl}
\end{algorithm}

%% file: listing/BeLevel-SSFL.tex
\begin{algorithm}[h!]
\small
\SetKwInOut{Input}{input}\SetKwInOut{Output}{return}
\Input{$K, T, \lambda, \Theta^{(0)}, \{\theta_i^{(0)}\}_{k\in[K]}, s\text{: number of local iteration}, \beta\text{: learning rate}$}
\For{$t=0,\ldots,T-1$}{
    Server randomly selects a subset of devices $S^{(t)}$\\
    Server sends the current global model $\Theta^{(t)}$ to $S^{(t)}$\\
    \For{device $k \in S^{(t)}$ in parallel}{
        \begin{tikzpicture} [remember picture, overlay]
            \node at (3pt, 2pt) (a) {};
            \node at (352pt,-93pt) (b) {};
            \node[line width=0pt, rounded corners=2pt, fill=blue!18, fill opacity=0.3, rectangle, fit=(a) (b)] (x) {};
            \node[anchor=south east,inner sep=1pt] at (x.south east) {\scriptsize$\clientOpt$};
        \end{tikzpicture}%
        Sample mini-batch $B_k$ from local dataset $D_k$, and do $s$ local iterations\\
        \tcc{Optimize the global parameter $\Theta$ locally}
        $Z_1, Z_2 \leftarrow f_{\Theta^{(t)}}(\Tau(B_k)), f_{\Theta^{(t)}}(\Tau(B_k))$\\
        $P_1, P_2 \leftarrow h_{\Theta^{(t)}}(Z_1), h_{\Theta^{(t)}}(Z_2)$\\
        $\Theta_k^{(t)} \leftarrow \Theta^{(t)} - \beta \nabla_{\Theta^{(t)}} \frac{\D(P_1, \widehat{Z_2}) + \D(P_2, \widehat{Z_1})}{2}$, where $\widehat{\cdot}$ stands for stop-gradient\\
        \tcc{Optimize the local parameter $\theta_k$}
        $z_1, z_2 \leftarrow f_{\theta_k}(\Tau(B_k)), f_{\theta_k}(\Tau(B_k))$\\
        $p_1, p_2 \leftarrow h_{\theta_k}(z_1), h_{\theta_k}(z_2)$\\
        $\theta_{k}\leftarrow\theta_{k}-\beta\nabla_{\theta_{k}}\left(\frac{\D(p1,\widehat{z_{2}})+\D(p2,\widehat{z_{1}})}{2}+\lambda\left\Vert \Theta^{(t)}-\theta_{k}\right\Vert _{2}^{2}\right)$\\
        Send $\Delta_k^{(t)}:=\Theta_k^{(t)}-\Theta^{(t)}$ back to server
    }
    \begin{tikzpicture} [remember picture, overlay]
        \node at (5pt, 5pt) (a) {};
        \node at (367pt, 0pt) (b) {};
        \node[line width=0pt, rounded corners=2pt, fill=green!18, fill opacity=0.3, rectangle, fit=(a) (b)] (x) {};
        \node[anchor=south east,inner sep=1pt] at (x.south east) {\scriptsize$\serverOpt$};
    \end{tikzpicture}%
    $\Theta^{(t+1)} \leftarrow \Theta^{(t)} + \sum_{k \in S^{(t)}} \frac{|D_k|}{|D|} \Delta_k^{(t)}$
}
\Output{$\{\theta_i\}_{i\in[n]}, \Theta^{(T)}$}
\caption{BiLevel-SSFL}
\label{alg:bilevel-ssfl}
\end{algorithm}

%% file: main_arxiv.bbl
\begin{thebibliography}{44}
\providecommand{\natexlab}[1]{#1}
\providecommand{\url}[1]{\texttt{#1}}
\expandafter\ifx\csname urlstyle\endcsname\relax
  \providecommand{\doi}[1]{doi: #1}\else
  \providecommand{\doi}{doi: \begingroup \urlstyle{rm}\Url}\fi

\bibitem[Kairouz et~al.(2021)Kairouz, McMahan, Avent, Bellet, Bennis, Bhagoji,
  Bonawitz, Charles, Cormode, Cummings, D'Oliveira, Rouayheb, Evans, Gardner,
  Garrett, Gasc{\'o}n, Ghazi, Gibbons, Gruteser, Harchaoui, He, He, Huo,
  Hutchinson, Hsu, Jaggi, Javidi, Joshi, Khodak, Konecn{\'y}, Korolova,
  Koushanfar, Koyejo, Lepoint, Liu, Mittal, Mohri, Nock, {\"O}zg{\"u}r, Pagh,
  Raykova, Qi, Ramage, Raskar, Song, Song, Stich, Sun, Suresh, Tram{\`e}r,
  Vepakomma, Wang, Xiong, Xu, Yang, Yu, Yu, and Zhao]{Kairouz2021AdvancesAO}
P.~Kairouz, H.~B. McMahan, B.~Avent, Aur{\'e}lien Bellet, M.~Bennis,
  A.~Bhagoji, Keith Bonawitz, Zachary~B. Charles, Graham Cormode, Rachel
  Cummings, Rafael G.~L. D'Oliveira, S.~Rouayheb, David Evans, Josh Gardner,
  Zachary Garrett, A.~Gasc{\'o}n, Badih Ghazi, Phillip~B. Gibbons, M.~Gruteser,
  Z.~Harchaoui, Chaoyang He, Lie He, Zhouyuan Huo, Ben Hutchinson, Justin Hsu,
  Martin Jaggi, T.~Javidi, Gauri Joshi, M.~Khodak, Jakub Konecn{\'y},
  Aleksandra Korolova, F.~Koushanfar, O.~Koyejo, Tancr{\`e}de Lepoint, Yang
  Liu, Prateek Mittal, M.~Mohri, R.~Nock, A.~{\"O}zg{\"u}r, R.~Pagh, Mariana
  Raykova, Hang Qi, D.~Ramage, R.~Raskar, D.~Song, Weikang Song, Sebastian~U.
  Stich, Ziteng Sun, A.~T. Suresh, Florian Tram{\`e}r, Praneeth Vepakomma,
  Jianyu Wang, Li~Xiong, Zheng Xu, Qiang Yang, F.~Yu, Han Yu, and Sen Zhao.
\newblock Advances and open problems in federated learning.
\newblock \emph{Found. Trends Mach. Learn.}, 14:\penalty0 1--210, 2021.

\bibitem[Wang et~al.(2021)Wang, Charles, Xu, Joshi, McMahan, Arcas,
  Al-Shedivat, Andrew, Avestimehr, Daly, Data, Diggavi, Eichner, Gadhikar,
  Garrett, Girgis, Hanzely, Hard, He, Horvath, Huo, Ingerman, Jaggi, Javidi,
  Kairouz, Kale, Karimireddy, Konecn{\'y}, Koyejo, Li, Liu, Mohri, Qi, Reddi,
  Richt{\'a}rik, Singhal, Smith, Soltanolkotabi, Song, Suresh, Stich,
  Talwalkar, Wang, Woodworth, Wu, Yu, Yuan, Zaheer, Zhang, Zhang, Zheng, Zhu,
  and Zhu]{Wang2021AFG}
Jianyu Wang, Zachary~B. Charles, Zheng Xu, Gauri Joshi, H.~B. McMahan, B.~A.~Y.
  Arcas, Maruan Al-Shedivat, Galen Andrew, S.~Avestimehr, Katharine Daly,
  Deepesh Data, S.~Diggavi, Hubert Eichner, Advait Gadhikar, Zachary Garrett,
  Antonious~M. Girgis, Filip Hanzely, Andrew Hard, Chaoyang He, Samuel Horvath,
  Zhouyuan Huo, A.~Ingerman, Martin Jaggi, T.~Javidi, P.~Kairouz, Satyen Kale,
  Sai Praneeth~Reddy Karimireddy, Jakub Konecn{\'y}, Sanmi Koyejo, Tian Li,
  Luyang Liu, M.~Mohri, Hang Qi, Sashank~J. Reddi, Peter Richt{\'a}rik,
  K.~Singhal, Virginia Smith, M.~Soltanolkotabi, Weikang Song, A.~T. Suresh,
  Sebastian~U. Stich, Ameet~S. Talwalkar, Hongyi Wang, Blake~E. Woodworth,
  Shanshan Wu, Felix~X. Yu, Honglin Yuan, M.~Zaheer, Mi~Zhang, Tong Zhang,
  Chunxiang Zheng, Chen Zhu, and Wennan Zhu.
\newblock A field guide to federated optimization.
\newblock \emph{ArXiv}, abs/2107.06917, 2021.

\bibitem[McMahan et~al.(2017)McMahan, Moore, Ramage, Hampson, and
  y~Arcas]{mcmahan2017communication}
Brendan McMahan, Eider Moore, Daniel Ramage, Seth Hampson, and Blaise~Aguera
  y~Arcas.
\newblock Communication-efficient learning of deep networks from decentralized
  data.
\newblock In \emph{Artificial Intelligence and Statistics}, pages 1273--1282,
  2017.

\bibitem[Li et~al.(2018)Li, Sahu, Zaheer, Sanjabi, Talwalkar, and
  Smith]{li2018federated}
Tian Li, Anit~Kumar Sahu, Manzil Zaheer, Maziar Sanjabi, Ameet Talwalkar, and
  Virginia Smith.
\newblock Federated optimization in heterogeneous networks.
\newblock \emph{arXiv preprint arXiv:1812.06127}, 2018.

\bibitem[Wang et~al.(2020)Wang, Liu, Liang, Joshi, and Poor]{wang2020tackling}
Jianyu Wang, Qinghua Liu, Hao Liang, Gauri Joshi, and H~Vincent Poor.
\newblock Tackling the objective inconsistency problem in heterogeneous
  federated optimization.
\newblock \emph{arXiv preprint arXiv:2007.07481}, 2020.

\bibitem[Reddi et~al.(2020)Reddi, Charles, Zaheer, Garrett, Rush,
  Kone{\v{c}}n{\`y}, Kumar, and McMahan]{reddi2020adaptive}
Sashank Reddi, Zachary Charles, Manzil Zaheer, Zachary Garrett, Keith Rush,
  Jakub Kone{\v{c}}n{\`y}, Sanjiv Kumar, and H~Brendan McMahan.
\newblock Adaptive federated optimization.
\newblock \emph{arXiv preprint arXiv:2003.00295}, 2020.

\bibitem[Liu et~al.(2020)Liu, Yuan, Zhao, Zheng, and Zheng]{liu2020rc}
Yi~Liu, Xingliang Yuan, Ruihui Zhao, Yifeng Zheng, and Yefeng Zheng.
\newblock Rc-ssfl: Towards robust and communication-efficient semi-supervised
  federated learning system.
\newblock \emph{arXiv preprint arXiv:2012.04432}, 2020.

\bibitem[Long et~al.(2020)Long, Che, Wang, Ye, Luo, Wu, Xiao, and
  Ma]{long2020fedsemi}
Zewei Long, Liwei Che, Yaqing Wang, Muchao Ye, Junyu Luo, Jinze Wu, Houping
  Xiao, and Fenglong Ma.
\newblock Fedsemi: An adaptive federated semi-supervised learning framework.
\newblock \emph{arXiv preprint arXiv:2012.03292}, 2020.

\bibitem[Itahara et~al.(2020)Itahara, Nishio, Koda, Morikura, and
  Yamamoto]{itahara2020distillation}
Sohei Itahara, Takayuki Nishio, Yusuke Koda, Masahiro Morikura, and Koji
  Yamamoto.
\newblock Distillation-based semi-supervised federated learning for
  communication-efficient collaborative training with non-iid private data.
\newblock \emph{arXiv preprint arXiv:2008.06180}, 2020.

\bibitem[Jeong et~al.(2020)Jeong, Yoon, Yang, and Hwang]{jeong2020federated}
Wonyong Jeong, Jaehong Yoon, Eunho Yang, and Sung~Ju Hwang.
\newblock Federated semi-supervised learning with inter-client consistency.
\newblock \emph{arXiv preprint arXiv:2006.12097}, 2020.

\bibitem[Liang et~al.(2021)Liang, Liu, Luo, He, Chen, and Yang]{liang2021self}
Xinle Liang, Yang Liu, Jiahuan Luo, Yuanqin He, Tianjian Chen, and Qiang Yang.
\newblock Self-supervised cross-silo federated neural architecture search.
\newblock \emph{arXiv preprint arXiv:2101.11896}, 2021.

\bibitem[Zhao et~al.(2020)Zhao, Liu, Li, Barnaghi, and Haddadi]{zhao2020semi}
Yuchen Zhao, Hanyang Liu, Honglin Li, Payam Barnaghi, and Hamed Haddadi.
\newblock Semi-supervised federated learning for activity recognition.
\newblock \emph{arXiv preprint arXiv:2011.00851}, 2020.

\bibitem[Zhang et~al.(2020{\natexlab{a}})Zhang, Kuang, You, Shen, Xiao, Zhang,
  Wu, Zhuang, and Li]{zhang2020federated}
Fengda Zhang, Kun Kuang, Zhaoyang You, Tao Shen, Jun Xiao, Yin Zhang, Chao Wu,
  Yueting Zhuang, and Xiaolin Li.
\newblock Federated unsupervised representation learning.
\newblock \emph{arXiv preprint arXiv:2010.08982}, 2020{\natexlab{a}}.

\bibitem[Zhang et~al.(2020{\natexlab{b}})Zhang, Yang, Yao, Yan, Gonzalez, and
  Mahoney]{zhang2020improving}
Zhengming Zhang, Yaoqing Yang, Zhewei Yao, Yujun Yan, Joseph~E Gonzalez, and
  Michael~W Mahoney.
\newblock Improving semi-supervised federated learning by reducing the gradient
  diversity of models.
\newblock \emph{arXiv preprint arXiv:2008.11364}, 2020{\natexlab{b}}.

\bibitem[Miyato et~al.(2018)Miyato, Maeda, Koyama, and
  Ishii]{miyato2018virtual}
Takeru Miyato, Shin-ichi Maeda, Masanori Koyama, and Shin Ishii.
\newblock Virtual adversarial training: a regularization method for supervised
  and semi-supervised learning.
\newblock \emph{IEEE transactions on pattern analysis and machine
  intelligence}, 41\penalty0 (8):\penalty0 1979--1993, 2018.

\bibitem[Lee(2013)]{Lee2013PseudoLabelT}
Dong-Hyun Lee.
\newblock Pseudo-label : The simple and efficient semi-supervised learning
  method for deep neural networks.
\newblock 2013.

\bibitem[Saeed et~al.(2020)Saeed, Salim, Ozcelebi, and
  Lukkien]{saeed2020federated}
Aaqib Saeed, Flora~D Salim, Tanir Ozcelebi, and Johan Lukkien.
\newblock Federated self-supervised learning of multisensor representations for
  embedded intelligence.
\newblock \emph{IEEE Internet of Things Journal}, 8\penalty0 (2):\penalty0
  1030--1040, 2020.

\bibitem[Bromley et~al.(1993)Bromley, Bentz, Bottou, Guyon, LeCun, Moore,
  S{\"a}ckinger, and Shah]{bromley1993signature}
Jane Bromley, James~W Bentz, L{\'e}on Bottou, Isabelle Guyon, Yann LeCun, Cliff
  Moore, Eduard S{\"a}ckinger, and Roopak Shah.
\newblock Signature verification using a “siamese” time delay neural
  network.
\newblock \emph{International Journal of Pattern Recognition and Artificial
  Intelligence}, 7\penalty0 (04):\penalty0 669--688, 1993.

\bibitem[Chen et~al.(2020)Chen, Kornblith, Norouzi, and Hinton]{chen2020simple}
Ting Chen, Simon Kornblith, Mohammad Norouzi, and Geoffrey Hinton.
\newblock A simple framework for contrastive learning of visual
  representations.
\newblock In \emph{International conference on machine learning}, pages
  1597--1607. PMLR, 2020.

\bibitem[Caron et~al.(2021)Caron, Misra, Mairal, Goyal, Bojanowski, and
  Joulin]{caron2021unsupervised}
Mathilde Caron, Ishan Misra, Julien Mairal, Priya Goyal, Piotr Bojanowski, and
  Armand Joulin.
\newblock Unsupervised learning of visual features by contrasting cluster
  assignments, 2021.

\bibitem[Grill et~al.(2020)Grill, Strub, Altch{\'e}, Tallec, Richemond,
  Buchatskaya, Doersch, Pires, Guo, Azar, et~al.]{grill2020bootstrap}
Jean-Bastien Grill, Florian Strub, Florent Altch{\'e}, Corentin Tallec,
  Pierre~H Richemond, Elena Buchatskaya, Carl Doersch, Bernardo~Avila Pires,
  Zhaohan~Daniel Guo, Mohammad~Gheshlaghi Azar, et~al.
\newblock Bootstrap your own latent: A new approach to self-supervised
  learning.
\newblock \emph{arXiv preprint arXiv:2006.07733}, 2020.

\bibitem[Chen and He(2020)]{chen2020exploring}
Xinlei Chen and Kaiming He.
\newblock Exploring simple siamese representation learning.
\newblock 2020.

\bibitem[Fallah et~al.(2020)Fallah, Mokhtari, and
  Ozdaglar]{fallah2020personalized}
Alireza Fallah, Aryan Mokhtari, and Asuman Ozdaglar.
\newblock Personalized federated learning: A meta-learning approach.
\newblock \emph{arXiv preprint arXiv:2002.07948}, 2020.

\bibitem[Li et~al.(2021)Li, Hu, Beirami, and Smith]{li_ditto_2021}
Tian Li, Shengyuan Hu, Ahmad Beirami, and Virginia Smith.
\newblock Ditto: Fair and robust federated learning through personalization.
\newblock 2021.
\newblock URL \url{http://arxiv.org/abs/2012.04221}.

\bibitem[Fallah et~al.()Fallah, Mokhtari, and
  Ozdaglar]{fallah_personalized_2020-2}
Alireza Fallah, Aryan Mokhtari, and Asuman Ozdaglar.
\newblock \emph{Personalized Federated Learning: A Meta-Learning Approach}.

\bibitem[Finn et~al.()Finn, Abbeel, and Levine]{finn_model-agnostic_2017}
Chelsea Finn, Pieter Abbeel, and Sergey Levine.
\newblock Model-agnostic meta-learning for fast adaptation of deep networks.
\newblock URL \url{http://arxiv.org/abs/1703.03400}.

\bibitem[Wu et~al.(2018)Wu, Xiong, Yu, and Lin]{wu2018unsupervised}
Zhirong Wu, Yuanjun Xiong, Stella Yu, and Dahua Lin.
\newblock Unsupervised feature learning via non-parametric instance-level
  discrimination, 2018.

\bibitem[Caron et~al.(2020)Caron, Misra, Mairal, Goyal, Bojanowski, and
  Joulin]{caron2020unsupervised}
Mathilde Caron, Ishan Misra, Julien Mairal, Priya Goyal, Piotr Bojanowski, and
  Armand Joulin.
\newblock Unsupervised learning of visual features by contrasting cluster
  assignments.
\newblock \emph{arXiv preprint arXiv:2006.09882}, 2020.

\bibitem[He et~al.(2020)He, Li, So, Zhang, Wang, Wang, Vepakomma, Singh, Qiu,
  Shen, Zhao, Kang, Liu, Raskar, Yang, Annavaram, and
  Avestimehr]{chaoyanghe2020fedml}
Chaoyang He, Songze Li, Jinhyun So, Mi~Zhang, Hongyi Wang, Xiaoyang Wang,
  Praneeth Vepakomma, Abhishek Singh, Hang Qiu, Li~Shen, Peilin Zhao, Yan Kang,
  Yang Liu, Ramesh Raskar, Qiang Yang, Murali Annavaram, and Salman Avestimehr.
\newblock Fedml: A research library and benchmark for federated machine
  learning.
\newblock \emph{arXiv preprint arXiv:2007.13518}, 2020.

\bibitem[Kairouz et~al.(2019)Kairouz, McMahan, Avent, Bellet, Bennis, Bhagoji,
  Bonawitz, Charles, Cormode, Cummings, et~al.]{kairouz2019advances}
Peter Kairouz, H~Brendan McMahan, Brendan Avent, Aur{\'e}lien Bellet, Mehdi
  Bennis, Arjun~Nitin Bhagoji, Keith Bonawitz, Zachary Charles, Graham Cormode,
  Rachel Cummings, et~al.
\newblock Advances and open problems in federated learning.
\newblock \emph{arXiv preprint arXiv:1912.04977}, 2019.

\bibitem[Hsu et~al.(2019)Hsu, Qi, and Brown]{hsu2019measuring}
Tzu-Ming~Harry Hsu, Hang Qi, and Matthew Brown.
\newblock Measuring the effects of non-identical data distribution for
  federated visual classification.
\newblock \emph{arXiv preprint arXiv:1909.06335}, 2019.

\bibitem[Dinh et~al.()Dinh, Tran, and Nguyen]{dinh_personalized_2020}
Canh~T. Dinh, Nguyen~H. Tran, and Tuan~Dung Nguyen.
\newblock Personalized federated learning with moreau envelopes.

\bibitem[Cuturi(2013)]{cuturi2013sinkhorn}
Marco Cuturi.
\newblock Sinkhorn distances: Lightspeed computation of optimal transportation
  distances, 2013.

\bibitem[Salimans and Kingma(2016)]{DBLP:journals/corr/SalimansK16}
Tim Salimans and Diederik~P. Kingma.
\newblock Weight normalization: {A} simple reparameterization to accelerate
  training of deep neural networks.
\newblock \emph{CoRR}, abs/1602.07868, 2016.
\newblock URL \url{http://arxiv.org/abs/1602.07868}.

\bibitem[Berthelot et~al.({\natexlab{a}})Berthelot, Carlini, Goodfellow,
  Papernot, Oliver, and Raffel]{berthelot_mixmatch_2019}
David Berthelot, Nicholas Carlini, Ian Goodfellow, Nicolas Papernot, Avital
  Oliver, and Colin~A. Raffel.
\newblock {MixMatch}: A holistic approach to semi-supervised learning.
\newblock In \emph{Neural Information Processing Systems}, pages 5049--5059,
  {\natexlab{a}}.

\bibitem[Laine and Aila()]{laine_temporal_2016}
Samuli Laine and Timo Aila.
\newblock Temporal ensembling for semi-supervised learning.

\bibitem[Sajjadi et~al.()Sajjadi, Javanmardi, and
  Tasdizen]{sajjadi_regularization_2016}
Mehdi Sajjadi, Mehran Javanmardi, and Tolga Tasdizen.
\newblock Regularization with stochastic transformations and perturbations for
  deep semi-supervised learning.
\newblock In \emph{Neural Information Processing Systems}, pages 1171--1179.

\bibitem[Berthelot et~al.({\natexlab{b}})Berthelot, Carlini, Cubuk, Kurakin,
  Sohn, Zhang, and Raffel]{berthelot_remixmatch_2020}
David Berthelot, Nicholas Carlini, Ekin~D. Cubuk, Alex Kurakin, Kihyuk Sohn,
  Han Zhang, and Colin Raffel.
\newblock {ReMixMatch}: Semi-supervised learning with distribution matching and
  augmentation anchoring.
\newblock In \emph{International Conference on Learning Representations},
  {\natexlab{b}}.

\bibitem[Hu et~al.()Hu, Yang, Hu, and Nevatia]{hu_simple_2021}
Zijian Hu, Zhengyu Yang, Xuefeng Hu, and Ram Nevatia.
\newblock {SimPLE}: Similar pseudo label exploitation for semi-supervised
  classification.
\newblock URL \url{http://arxiv.org/abs/2103.16725}.

\bibitem[Miyato et~al.()Miyato, Maeda, Koyama, and Ishii]{miyato_virtual_2019}
Takeru Miyato, Shin-Ichi Maeda, Masanori Koyama, and Shin Ishii.
\newblock Virtual adversarial training: A regularization method for supervised
  and semi-supervised learning.
\newblock 41\penalty0 (8):\penalty0 1979--1993.
\newblock \doi{10.1109/TPAMI.2018.2858821}.

\bibitem[{Brendan McMahan} et~al.(2016){Brendan McMahan}, {Moore}, {Ramage},
  {Hampson}, and {Ag{\"u}era y Arcas}]{brendan2016}
H.~{Brendan McMahan}, Eider {Moore}, Daniel {Ramage}, Seth {Hampson}, and
  Blaise {Ag{\"u}era y Arcas}.
\newblock {Communication-Efficient Learning of Deep Networks from Decentralized
  Data}.
\newblock \emph{arXiv e-prints}, art. arXiv:1602.05629, February 2016.

\bibitem[Tan and Le(2019)]{tan2019efficientnet}
Mingxing Tan and Quoc Le.
\newblock Efficientnet: Rethinking model scaling for convolutional neural
  networks.
\newblock In \emph{International Conference on Machine Learning}, pages
  6105--6114. PMLR, 2019.

\bibitem[Howard et~al.(2017)Howard, Zhu, Chen, Kalenichenko, Wang, Weyand,
  Andreetto, and Adam]{howard2017mobilenets}
Andrew~G Howard, Menglong Zhu, Bo~Chen, Dmitry Kalenichenko, Weijun Wang,
  Tobias Weyand, Marco Andreetto, and Hartwig Adam.
\newblock Mobilenets: Efficient convolutional neural networks for mobile vision
  applications.
\newblock \emph{arXiv preprint arXiv:1704.04861}, 2017.

\bibitem[Cai et~al.(2020)Cai, Gan, Zhu, and Han]{cai2020tinytl}
Han Cai, Chuang Gan, Ligeng Zhu, and Song Han.
\newblock Tinytl: Reduce memory, not parameters for efficient on-device
  learning.
\newblock \emph{Advances in Neural Information Processing Systems}, 33, 2020.

\end{thebibliography}
